\documentclass[accepted]{uai2024} 
                        
\usepackage[american]{babel}

\usepackage{natbib} 
\usepackage{mathtools} 
\usepackage{booktabs} 
\usepackage{tikz} 

\usepackage{hyperref}
\usepackage{url}
\usepackage{multirow}
\usepackage{graphicx}
\usepackage[utf8]{inputenc} 
\usepackage[T1]{fontenc}    
\usepackage{url}            
\usepackage{booktabs}       
\usepackage{amsfonts}       
\usepackage{nicefrac}       
\usepackage{microtype}      

\usepackage{multirow}
\usepackage{graphicx}
\usepackage{amsmath}
\usepackage{amssymb}
\usepackage{subcaption}
\usepackage{balance}
\usepackage{diagbox}
\usepackage{algorithm} 
\usepackage[noend]{algpseudocode}
\usepackage{balance}

\usepackage{amsmath,amsfonts,bm}

\newcommand{\td}[1]{\Tilde{#1}}

\def\eqref#1{equation~\ref{#1}}

\def\1{\bm{1}}

\def\rvx{{\mathbf{x}}}
\def\rvy{{\mathbf{y}}}

\DeclareMathAlphabet{\mathsfit}{\encodingdefault}{\sfdefault}{m}{sl}
\SetMathAlphabet{\mathsfit}{bold}{\encodingdefault}{\sfdefault}{bx}{n}

\def\gL{{\mathcal{L}}}

\def\gN{{\mathcal{N}}}

\def\tdy{{\Tilde{y}}}

\newcommand{\E}{\mathbb{E}}

\title{Bayesian Pseudo-Coresets via Contrastive Divergence}

\author[1]{Piyush~Tiwary}
\author[1]{Kumar~Shubham}
\author[1]{Vivek~V.~Kashyap}
\author[1]{Prathosh~A.P.}
\affil[1]{%
    Department of Electrical Communuication Engineering\\
    
    Indian Institute of Science\\
    Bengaluru, Karnataka 560012, India
}
  
\begin{document}
\maketitle

\begin{abstract}
  Bayesian methods provide an elegant framework for estimating parameter posteriors and quantification of uncertainty associated with probabilistic models. However, they often suffer from slow inference times. To address this challenge, Bayesian Pseudo-Coresets (BPC) have emerged as a promising solution. BPC methods aim to create a small synthetic dataset, known as pseudo-coresets, that approximates the posterior inference achieved with the original dataset. This approximation is achieved by optimizing a divergence measure between the true posterior and the pseudo-coreset posterior.
    Various divergence measures have been proposed for constructing pseudo-coresets, with forward Kullback-Leibler (KL) divergence being the most successful. However, using forward KL divergence necessitates sampling from the pseudo-coreset posterior, often accomplished through approximate Gaussian variational distributions. Alternatively, one could employ Markov Chain Monte Carlo (MCMC) methods for sampling, but this becomes challenging in high-dimensional parameter spaces due to slow mixing.
    In this study, we introduce a novel approach for constructing pseudo-coresets by utilizing contrastive divergence. Importantly, optimizing contrastive divergence eliminates the need for approximations in the pseudo-coreset construction process. Furthermore, it enables the use of finite-step MCMC methods, alleviating the requirement for extensive mixing to reach a stationary distribution.
    To validate our method's effectiveness, we conduct extensive experiments on multiple datasets, demonstrating its superiority over existing BPC techniques. 
    Our implementation is available at \href{https://github.com/backpropagator/BPC-CD}{https://github.com/backpropagator/BPC-CD}
\end{abstract}

\section{Introduction}\label{sec:intro}
In recent years, contemporary deep learning models have demonstrated exceptional effectiveness in a wide array of applications, spanning computer vision, natural language processing, and speech analysis~\citep{krizhevsky2017imagenet,devlin2018bert,amodei2016deep,he2016deep,dosovitskiy2020image,radford2021learning}. 
Conventional deep learning methods rely on one-time training of models providing point estimates~\citep{szegedy2013intriguing}. These point estimates are prone to overfitting and often provide overconfident or under-confident outputs~\citep{gawlikowski2023survey,kabir2018neural}. This prohibits the use of deep learning models in critical applications such as medical, finance, etc~\citep{ker2017deep,cavalcante2016computational}.
Bayesian methods furnish a systematic framework for parameter estimation and quantification of associated uncertainty. Bayesian inference entails sampling from parameter posterior distributions using Markov Chain Monte Carlo (MCMC) techniques~\citep{robert1999monte,robert2011short}. However, conducting inference based on parameter posterior conditioned on the entire dataset is computationally demanding, particularly as the dataset size, denoted by $N$, increases. The computational complexity of MCMC methods scales with $N$ as $\Theta(NS)$, where $S$ denotes the number of samples~\citep{campbell_tamara_greedy_2018}. This complexity becomes prohibitively high for large $N$. To mitigate this, one often resorts to using a random subset of $M \ll N$ data points for likelihood computation at each iteration~\citep{bardenet2017markov,korattikara2014austerity,maclaurin2014firefly,welling2011bayesian,ahn2012bayesian,bierkens2019zig,pollock2020quasi}. However, such approximations introduce errors and lead to slow mixing of Markov chains~\citep{johndrow2020no,nagapetyan2017true,betancourt2015fundamental}.

Bayesian coresets~\citep{huggins_tamara_logistic} were introduced to solve the aforementioned problem. Particularly,~\citet{huggins_tamara_logistic} proposed to select a subset of original dataset (also called a coreset) that uniformly approximates the  log-likelihood of the original dataset. These coresets are significantly smaller in size than the original dataset, leading to vastly improved sampling efficiency. Further,~\cite{campbell_variational_bayesiancoreset} proposed to identify the coreset by minimizing Kullback-Leibler (KL) divergence between full data posterior and coreset posterior. However, most of such methods do not scale with data dimension~\citep{BPC_OG}. Particularly, the KL-divergence between the (optimal) coreset posterior and true posterior increases with data dimension. Meaning that for large data dimension, even with the optimal coresets, the KL-divergence is far from optima (zero), implying that the true posterior is not approximated correctly. However, recently few methods have tried to overcome the scalability issue by using lightweight coresets~\citep{bachem2018scalable}.

The Bayesian Pseudo-Coreset (BPC)~\citep{BPC_OG} approach, as a distinct category of methods, has been proposed to \textit{synthesise} a smaller dataset from the original one, as opposed to selecting a subset, which is the case with coreset methods. The fundamental idea of BPC involves solving an optimization problem over the data space, leading to creation of a `pseudo' dataset that appropriately approximates the true posterior. The said optimization problem pertains to minimization of a divergence metric between the true posterior and pseudo-coreset posterior. 
Such a framework removes the constraint for the pseudo-coresets to be a subset of the original dataset. This additional degree of freedom aids in better optimization of the divergence measure.
Further, in addition to better approximation of true posterior, pseudo-coreset also come with privacy benefits. Specifically, since pseudo-coresets are not part of original dataset, one can outsource these pseudo-coreset without revealing the original dataset for an user to run inference. \cite{BPC_OG} also provided theoretical guarantees showing pseudo-coresets are differentially private.

Recently,~\cite{BPC} analyzed the BPC construction under different divergence measures such as reverse-KL and Wasserstein divergence. Their analysis revealed that BPC methods under different divergence measures are equivalent to their non-bayesian counterparts. These non-bayesian frameworks are often referred to as `Dataset Condensation' or `Dataset Distillation'~\citep{GM,DM,MTT,CAFE,KIP}. In particular,~\cite{BPC} showed that minimization of reverse-KL is equivalent to gradient matching~\citep{GM} and minimization of wasserstein measure is equivalent to matching training trajectory~\citep{MTT}. 
They also proposed to use forward-KL for better pseudo-coreset construction due to its ability to capture the support of the distribution, contrasting with reverse-KL, which tends to focus on the distribution's modes. However, computing the gradient of forward-KL requires sampling from the intractable pseudo-coreset posterior. While this can be achieved using MCMC methods, the extensive mixing time of MCMC in high-dimensional parameter spaces renders this approach impractical. As a remedy, Gaussian variational approximation around SGD solutions was employed to simplify and expedite the sampling process.  However, the quality of such approximation is unknown and remains a matter of concern.  
Here, we note that the definition of contrastive divergence as presented in \citet{BPC} does not align with the original definition in \citet{CD}. Specifically, while \citet{BPC} uses forward-KL divergence to arrive at an objective referred to as the contrastive divergence (Section 3.3 of \citet{BPC}), the original definition (in \citet{CD})  involves difference between two forward-KL divergences. In this work, we resort to the original definition of contrastive divergence as noted in \citet{CD}.


In our current work, we propose a novel approach: using contrastive divergence instead of forward-KL divergence for pseudo-coreset learning. This has two advantages: (1) It eliminates the need for approximating the pseudo-coreset posterior, enabling the straightforward use of MCMC methods, (2) The Markov chain used in this approach does not require extensive mixing to reach a stationary distribution; only a finite number of steps is needed. These advantages effectively address the challenges associated with using forward-KL divergence. Furthermore, our rigorous experiments demonstrate that our proposed method significantly outperforms previous state-of-the-art BPC methods, thereby confirming that the pseudo-coreset posterior using contrastive divergence better approximates the true posterior. Our contributions can be summarized as follows:
\begin{itemize}
    \item We propose a new framework for the construction of Bayesian Pseudo-Coreset  using contrastive divergence.
    \item The proposed method avoids any approximation of pseudo-coreset posterior and facilitates the use of finite step MCMC methods during learning phase.
    \item Extensive experimentation reveals that our method surpasses state-of-the-art BPC methods by substantial margins, affirming the better approximation of the true posterior using contrastive divergence.
\end{itemize}

\section{Related Work}
\label{sec:related_works}

\subsection{Bayesian Inference and Optimization}
The objective of Bayesian methods is the model the parameter posterior distribution of a probabilistic model. However, apart from some simple models, the exact posterior distributions are generally intractable ~\citep{campbell_tamara_hilbert_2019}. In such scenarios, one often relies on inference techniques like MCMC methods~\citep{robert1999monte,robert2011short} and Variational Inference (VI)~\citep{jordan1998introduction,wainwright2008graphical}. Historically, these inference techniques require model-specific tuning based on the path-length parameters, step size~\citep{neal2011handbook}, and the choice of the variational families~\citep{jaakkola1997variational,jordan1999introduction}. Recent methods~\citep{ranganath2014black,kucukelbir2017automatic,hoffman2014no} have circumvented these issues by introducing a black box approach that requires only basic specifications about the model. For instance, the traditional variational inference methods~\citep{jaakkola1997variational,jordan1999introduction} relied on closed form gradients of the model~\citep{ranganath2014black} and an approximate distribution for the posterior of the data. \citet{ranganath2014black,baydin2018automatic,kucukelbir2017automatic} addressed these issues by employing standard transformation over a multivariate Gaussian distribution and used automatic differentiation techniques to calculate the associated gradients. Similarly, for MCMC methods like Hamiltonian Monte Carlo (HMC)~\citep{neal2011handbook} traditional practices involved manually tuning of parameters like step size and path length to achieve accurate posterior estimation. \citet{hoffman2014no} addressed this challenge by automatically estimating both of these parameters.   


In many modern applications, these methods are required to scale with the size of the datasets. The standard MCMC algorithms are computationally expensive for large datasets, and the sampling process scales linearly with the data size. Recent works~\citep{bardenet2017markov,korattikara2014austerity,maclaurin2014firefly,welling2011bayesian,ahn2012bayesian,bierkens2019zig,pollock2020quasi}, have tried to mitigate the computational cost associated with inference models by considering only a random subset of data points during MCMC iterations. One of the initial studies in this direction has been conducted by~\citet{welling2011bayesian} where the authors proposed to use stochastic gradient langevin dynamics (SGLD). This iterative learning algorithm utilizes mini-batches of dataset for Bayesian inference. However, unlike other MCMC methods, their approach often leads to a slow mixing rate. \citet{ahn2012bayesian} addressed this issue by sampling from the Gaussian approximation of posterior for a high mixing rate and mimicking the behavior of SGLD using a pre-conditioner matrix for a slow mixing rate. However, \citet{korattikara2014austerity,bardenet2014towards} have shown that such a sampling approach often leads to a stationary distribution that can have bounded errors under strong conditions of rapid mixing~\citep{maclaurin2014firefly}.
In contrast, they proposed a new accept/reject strategy to select a subset of the dataset for Bayesian inference. On a similar line,~\citet{maclaurin2014firefly} proposed to use a collection of Bernoulli latent variables to select a subset of the dataset for likelihood estimation.  \citet{bierkens2019zig,pollock2020quasi}  have further proposed to use a zig-zag process and quasi-stationary distribution along with the subsampling approaches for bayesian inference.


\subsection{Bayesian Coresets}
\label{sec:bayesian-coreset}
Bayesian coresets~\citep{huggins_tamara_logistic,campbell_tamara_greedy_2018,campbell_variational_bayesiancoreset,campbell_tamara_hilbert_2019,zhang_bayesiancoreset,naik2022fast,chen2022bayesian} present an alternative strategy to address aforementioned challenges by selecting a small weighted subset of the original dataset which can closely approximate the posterior of the full dataset~\citep{zhang_bayesiancoreset,huggins_tamara_logistic}. 
The idea was introduced in~\citet{huggins_tamara_logistic}, where a weighted subset of original data was selected to approximate the log-likelihood of the entire dataset up to some multiplicative error over the parameter space. However, the subset produced by such a technique underestimates the posterior distribution and can result in large approximation errors for some models regardless of the coreset size. \citet{campbell_tamara_greedy_2018} addressed this issue using greedy iterative geodesic ascent (GIGA), that optimally scales the log-likelihood of the coreset to better approximate the entire log-likelihood of the dataset. It further provided a uniform bounded error for all the models. To further enhance the scalability, \citet{campbell_tamara_hilbert_2019} tackled the model and data-specific assumptions made in prior work regarding coreset construction. They constructed Bayesian coreset by solving a sparse vector sum based approximation using frank-wolfe~\citep{frank1956algorithm} based solvers. Recent works~\citep{zhang_bayesiancoreset,naik2022fast,chen2022bayesian} have focused on improving the speed of coreset construction using accelerated optimization methods, quasi-newton refinement, and sparse-hamiltonian flows. However, since the KL divergence between the posteriors of the optimal coreset and the original dataset increases with the data dimensionality~\citep{BPC_OG}, these methods do not easily scale up in high-dimensions. 

\subsection{Bayesian Pseudo-Coreset}
\label{sec:BPC}
\citet{BPC_OG} proposed to use a collection of synthetic data to scale the Bayesian inference to high dimensional datasets. Particularly, they frame the problem as divergence minimization between the posteriors associated with the synthetic and the original dataset. The synthetic set generated through this technique is called `Bayesian Pseudo-Coreset' (BPC). Compared to Bayesian coresets, these methods scale more efficiently with data dimensions and yield a more accurate posterior approximation.

\citet{BPC_OG} formalized the given problem by minimizing the reverse-KL divergence between the posterior of original data and the posterior of synthetic data. On similar lines,~\citet{BPC} demonstrated that other divergence metrics, such as Wasserstein distance and forward-KL divergence, can be used to generate pseudo-coreset. In contrast to reverse-KL, which primarily focuses on the modes of the distributions, forward-KL provides a mechanism to better capture the support of the posterior distribution. To efficiently calculate the forward-KL divergence~\citet{BPC} used a Gaussian variational approximation of the posterior distribution. However, the quality of such an approximation and its impact on the overall performance of the pseudo-coreset is unknown. Further, computing the gradient of forward-KL requires sampling from an intractable posterior of pseudo-coreset using MCMC methods, which is not straightforward in practice.

\subsection{Coresets and Dataset Condensation}
\label{sec:coreset}

While Bayesian coreset focuses on selecting data points to facilitate Bayesian inference, coreset selection strategies have been proposed for other algorithms like geometric approximation~\citep{agarwal2005geometric}, mixture models~\citep{feldman2011scalable}, K-means clustering~\citep{feldman2011unified,feldman2020turning,bachem2016approximate} and DP means~\citep{bachem2015coresets}. Similarly, for deep learning models, \citet{mirzasoleiman2020coresets,killamsetty2021grad,killamsetty2021glister} have introduced subset selection techniques that leverage gradient matching and meta-learning algorithms. Recent works~\citep{welling2009herding, castro_herding, icarl_herding,scail_herding,sener2017active,farahani2009facility}, have further proposed strategies to choose a representative and diverse set of samples from the original dataset. These methods aim to create a generic subset by removing redundant data points. Herding-based coreset methods~\citep{welling2009herding, castro_herding, icarl_herding,scail_herding} select such samples by minimizing the distance between the feature centroids of the coreset, and the original dataset. While K-center-based coreset techniques~\citep{sener2017active,farahani2009facility,guo2022deepcore} pick the most diverse and representative samples by optimizing a submodular function~\citep{farahani2009facility}. Contrary to K-center and herding-based coreset selection methods, forgetting-based coreset~\citep{Forgetting} removes the easily forgettable samples from the training dataset. 

Rather than selecting a subset of data points from the training set, dataset condensation methods aim to generate a synthetic set that emulates the characteristics of the original dataset. For example, in gradient based dataset condensation techniques~\citep{GM,DD_Survey, DC_CS,jiang2022delving} the synthetic samples are generated by aligning the gradients of a model trained using original and synthetic datasets. Similarly, meta-learning based methods~\citep{DD,Rem_Past,KIP,RFAD,FREPO} generate these synthetic samples by matching the validation performance of a model trained using the entire dataset with the performance of a model trained using the synthetic set. \citet{MTT,DD_PP,du2022minimizing} propose generating the synthetic dataset using long-horizon trajectories, ensuring that the models learn similar trajectories during optimization. While distribution matching methods~\citep{DM,CAFE,zhao2022synthesizing,Improve_DM} generate a condensed synthetic set with a similar feature distribution as the original dataset. Recent works~\citep{liu2023dream, zhang2022accelerating,cazenavette2023generalizing} have further focused on improving the performance and computational complexity of existing dataset condensation techniques by using representative samples from the training set, model augmentation techniques, and generative model for learning the synthetic set. While dataset condensation and BPC might seem to do the same, they are fundamentally different from each other. One can look at Bayesian Pseudo-Coresets as bayesian counterparts of dataset condensation methods. We provide detailed account of differences between dataset condensation methods and BPC methods in Appendix.

\section{Proposed Methodology} 
\label{sec:proposed-methodology}

\subsection{Bayesian Pseudo-Coresets}
\label{sec:BPC-BG}
Consider a  dataset 
$(\rvx,\rvy) = \{(\rvx_i,y_i)\}^{n}_{i=1}$ consisting of $n$ data points. 
Now consider a synthetic (learnable) dataset $(\td{\rvx},\td{\rvy}) = \{\td{\rvx}_i,\tdy_i\}_{i=1}^{m}$ such that $\rvy$ and $\td{\rvy}$ share the same label space and $m \ll n$.
Let, $\theta \in \Theta$ be the parameter of a discriminative / classification model. Then the parameter posteriors corresponding to original and synthetic data, $\pi(\theta|\rvx)$ and $\pi(\theta|\td{\rvx})$ are given by
\begin{align}
    \pi_{\rvx}  \triangleq \pi(\theta|\rvx) & = \frac{\pi_0(\theta)}{Z(\rvx)} \exp{\left(\sum\limits_{i=1}^{n} \log \pi(y_i|\rvx_i,\theta)\right)} \label{eq:true-posterior}\\
    \pi_{\td{\rvx}}  \triangleq \pi(\theta|\td{\rvx}) &= \frac{\pi_0(\theta)}{Z(\td{\rvx})} \exp{\left(\sum\limits_{i=1}^{m} \log \pi(\td{y}_i|\td{\rvx}_i,\theta)\right)}\\
    &= \frac{\pi_0(\theta)}{Z(\td{\rvx})} \exp{\left(-E(\td{\rvx},\theta)\right)} \label{eq:pc-posterior}
\end{align}
where, 
\begin{align}
    Z(\rvx) &= \int_{\Theta} \pi_0(\theta)\exp{\left(\sum_{i=1}^{n} \log \pi(y_i|\rvx_i,\theta)\right)} d\theta\\  Z(\td{\rvx}) &= \int_{\Theta} \pi_0(\theta)\exp{\left(-E(\td{\rvx},\theta)\right)} d\theta
\end{align}
are appropriate normalizing constants. Here, $\pi_0(\theta)$ is the prior distribution and $E(\td{\rvx},\theta) = -\sum\limits_{i=1}^{m} \log \pi(\td{y}_i|\td{\rvx}_i,\theta) $ is the sum of negative log-likelihoods which can be treated as a generic potential or energy function. Since $n$ is often very large, the posterior estimation using $\pi_\rvx$ is computationally expensive and infeasible. However, an appropriate approximation such as $\pi_{\td{\rvx}}$ where $m \ll n$, allows one to overcome this hurdle. In particular, this approximation is carried out by solving the following optimization problem:
\begin{align}
    \td{\rvx}^* = \underset{\td{\rvx}}{\arg\min}\hspace{2mm} D\left( \pi_{\rvx}, \pi_{\td{\rvx}} \right)
\end{align}
where, $D(\cdot,\cdot)$ is a divergence measure between two distributions. Recently, \cite{BPC} showed the results for above optimization problem under different divergence metrics. Specifically, they analyzed the results with reverse-KL and wasserstein divergence; consequently drawing equivalence with dataset condensation methods like gradient matching~\citep{GM} and MTT~\citep{MTT}. Further, they propose an alternative solution by using forward-KL divergence as it encourages a model to cover the entire target distribution in contrast to reverse-KL which encourages mode capturing models. The gradient of the forward-KL divergence, as derived in~\cite{BPC}, is expressed as follows:
\begin{align}
    \nabla_{\td{\rvx}} D_{KL}\left( \pi_{\rvx} || \pi_{\td{\rvx}} \right) = \E_{\pi_{\td{\rvx}}}\left[ -\nabla_{\td{\rvx}}  E(\td{\rvx},\theta)  \right] + \nabla_{\td{\rvx}} \E_{\pi_{\rvx}}\left[ E(\td{\rvx},\theta) \right]
\label{eq:fkl-loss}
\end{align}
This gradient computation necessitates the calculation of expectations with respect to the probability distributions $\pi_{\rvx}$ and $\pi_{\td{\rvx}}$. However, the presence of intractable partition functions ($Z(\rvx)$ and $Z(\td{\rvx})$) poses challenges in efficiently sampling from these posterior distributions. 
One can resort to MCMC methods such as langevin dynamics or hamiltonian monte-carlo for sampling, however, due to large dimension of $\Theta$-space, the mixing-time of these methods is very large and in-efficient in practice. 
To overcome this issue,~\cite{BPC} employs gaussian variational approximations for these posteriors, rendering the sampling process computationally feasible. Specifically, gaussian distributions are used, centered around parameters obtained from Stochastic Gradient Descent (SGD) trajectories of $\rvx$ and $\td{\rvx}$ (cf.~\citep{BPC} for details). 
\par In practice, since $m$ (number of samples in pseudo-coreset) is generally very small, the SGD trajectories of $\td{\rvx}$ might overfit, leading to erroneous approximations. 
Hence, it can be seen that there is a clear trade-off between `posterior approximation quality' and `computational efficiency' in the previous methods (cf. Table~\ref{tab:mcmc-comp} for quantitative numbers).
Therefore, it is preferable to bypass such approximations and sample directly from the exact posteriors. In this work, we propose to work with contrastive divergence~\citep{CD} instead of forward-KL to construct the pseudo-coreset. Specifically, using contrastive divergence leads to a loss objective where  $\pi_{\td{\rvx}}$ can be used as it is without any approximation. The key idea behind this is that instead of minimizing forward-KL, contrastive divergence minimizes difference between two forward-KL terms, that results in cancellation of expectation w.r.t $\pi_{\td{\rvx}}$ allowing us to circumvent this approximation. We describe this in detail in next section.

\subsection{Contrastive Divergence for BPC}
\label{sec:CD-for-BPC}
As mentioned earlier, we propose to work with contrastive divergence instead of forward-KL for construction of pseudo-coresets. The concept of contrastive divergence was initially introduced by seminal work in~\cite{CD}.The central premise behind contrastive divergence hinges on a straightforward insight: whereas minimizing forward KL divergence necessitates a term that involves sampling from $\pi_{\td{\rvx}}$, minimizing the difference between two forward KL divergences leads to the nullification of this term. More explicitly, the contrastive divergence is defined as:
\begin{align}
    \gL_{CD} = D_{KL}(\pi_{\rvx} || \pi_{\td{\rvx}}) - D_{KL}(\Pi_{E}^{k}\pi_{\rvx} || \pi_{\td{\rvx}})
\label{eq:final_cd}
\end{align}
where, $\Pi_{E}^{k}(\cdot)$ is an MCMC transition kernel for $\pi_{\td{\rvx}}$ and $\Pi_{E}^{k}\pi_{\rvx}$ represents $k$ sequential MCMC transitions starting from $\pi_{\rvx}$. 
Here, we use transition kernel in the context of Markov process. Particularly, a one-step transition kernel is a map that takes a state as input and generates the next state after one step. Similarly, a $k$-step transition kernel takes a state as input and generates the state after $k$-steps. This is akin to the role of transition matrix in markov chains with finite states. 
In context of the proposed method, the $k$-step transition kernel takes a state sampled from $\pi_\rvx$, and generates a state after $k$-steps. Additionally, we note that Eq.~\ref{eq:final_cd} is minimized to zero only if $\pi_\rvx$ = $\pi_{\tilde{\rvx}}$. This is a well known result noted in \citet{CD} (cf Page 4 of \citet{CD}).

For brevity, let us denote $\Bar{\pi}_{\rvx}$ as $\Pi_{E}^{k}\pi_{\rvx}$. As shown in~\cite{CD}, the gradient of the above objective is approximately given by~\footnote{Please note that the signs in this expression are opposite to those presented in \citet{BPC}, primarily due to differences in the treatment of the energy function. In \citet{BPC}, the energy function (referenced as Eq. 2 in their paper) is considered positive, whereas in our work, we adopt a convention where the energy function (as represented in Eq.~\ref{eq:pc-posterior} of our paper) is negative. This choice is made for convenience, aligning with the convention in physics literature where lower energy states are typically considered stable.}:
\begin{align}
    \nabla_{\td{\rvx}} \gL_{CD} = \E_{\pi_{\rvx}}\left[\nabla_{\td{\rvx}} E(\td{\rvx},\theta)\right] - \E_{\Bar{\pi}_{\rvx}}\left[\nabla_{\td{\rvx}} E(\td{\rvx},\theta)\right]
\end{align}
It is worth noting that the gradient estimation in the above equation does not necessitate sampling from $\pi_{\td{\rvx}}$. Instead, it calls for sampling from $\pi_{\rvx}$ and $\Bar{\pi}_{\rvx}$. In this context, we can employ a variational posterior to approximate  $\pi_{\rvx}$ and use MCMC sampling techniques (e.g. langevin dynamics~\citep{LD}) starting from $\pi_{\rvx}$ to sample from $\Bar{\pi}_{\rvx}$. Notably, unlike in Eq.~\ref{eq:fkl-loss}, the MCMC sampling utilized here only needs to run for finite $k$ steps, alleviating the requirement for substantial Markov chain mixing.

In particular, we use gaussian variational posterior ($q_{\rvx}$) to approximate $\pi_{\rvx}$. Then, a $k$-step MCMC starting from $q_{\rvx}$ should be used as a variational substitute for $\Bar{\pi}_{\rvx}$:
\begin{align}
    q_{\rvx}(\theta) = \gN(\theta; \theta_{\rvx}, \Sigma_{\rvx}),\hspace{4mm} \Bar{q}_{\rvx}(\theta) = \Pi_{E}^{k}~ q_{\rvx}(\theta)
\label{eq:var_approx}
\end{align}
where, $\theta_{\rvx}$ is the MAP solution computed for $\rvx$.
Here, one can note that making an approximation for $\pi_{\rvx}$ is enough unlike previous methods where additional approximations for $\pi_{\td{\rvx}}$ is also required. Hence, the final gradient estimate is obtained as
\begin{align}
    &\nabla_{\td{\rvx}} \gL_{CD} \approx \E_{q_{\rvx}}\left[\nabla_{\td{\rvx}} E(\td{\rvx},\theta)\right] - \E_{\Bar{q}_{\rvx}}\left[\nabla_{\td{\rvx}} E(\td{\rvx},\theta)\right] \\
    &\approx \nabla_{\td{\rvx}}\frac{1}{N} \sum_{j=1}^{N} \left[ E\left(\td{\rvx}, \theta_{\rvx} + \Sigma_{\rvx}^{1/2}\varepsilon_{\rvx}^{(j)}\right) - E\left(\td{\rvx}, \texttt{sg}\left(\Bar{\theta}^{(j)}\right)\right) \right] \label{eq:final-loss}
\end{align}
where, $\texttt{sg}(\cdot)$ denotes stop-gradient operator, $\varepsilon_{\rvx}^{(j)} \sim \gN(0,I)$ and $\Bar{\theta}^{(j)}$ is obtained via running $k$-step MCMC starting from $\left(\theta_{\rvx} + \Sigma_{\rvx}^{1/2}\varepsilon_{\rvx}^{(j)}\right)$. 
\par Here, we note that one can theoretically learn $\Sigma_\rvx$ by treating it as a learnable parameter. However, this would  require gradient of determinant of the covariance matrix. Given the high dimensional parameter space, this operation would lead to computational inefficiencies. Hence, we treat it as a hyperparameter and keep it fixed. 
\par Further, for computational efficiency, we assess the parameter posterior with $\rvx$ using expert trajectories similar to~\citet{BPC}. In essence, expert trajectories represent sequences of parameters obtained while training a model on the dataset ($\rvx$, $\rvy$). Each of these sequences is termed as `parameter trajectory,' and the collection of these trajectories, acquired through various training instances, is known as `expert trajectories.' This  eliminates the need to compute MAP solutions for $\rvx$ ($\theta_\rvx$) at each training step. During training, we randomly pick a parameter from these trajectories to calculate the objective function.

\section{Experiments and results}
\label{sec:experiments}
\subsection{Evaluation Details}

We evaluate our method both quantitatively and qualitatively on several BPC-benchmark datasets with different compression ratios, i.e., the number of images generated per class (ipc).
In particular, we perform our experiments on six different datasets, namely, CIFAR10~\citep{CIFAR10}, SVHN~\citep{SVHN}, MNIST~\citep{MNIST}, FashionMNIST~\citep{FashionMNIST}, CIFAR100~\citep{CIFAR10} and Tiny Imagenet (T-Imagenet)~\citep{tinyimagenet}. All the experiments perform multi-class classification tasks with ipc=1, 10, and 50 which is in line with previous baselines.We employ Langevin dynamics~\citep{neal2011mcmc,teh2003energy} during training as well as inference and report accuracy (Acc) and negative log-likelihood (NLL) with respect to the ground truth labels. For our primary experiments, we use a CNN architecture (ConvNet) exactly as described in the previous works \citep{kim2022dataset,BPC_OG,cazenavette2023generalizing} (cf. Appendix for details) for a fair comparison. 

Further, we assess the robustness of the BPC methods on  out-of-distribution dataset and against adversarial attacks in Section~\ref{sec:ood-results}. We also examine the cross-architecture performance of the proposed method in Section~\ref{sec:cross-arch}. Next, since bayesian methods are often sensitive to the number of parameters being sampled from the posterior, we observe the effect of number of parameters  on the proposed method and compare it with previous BPC baselines in Section~\ref{sec:diff-arch}. Further, we provide an ablation study of the proposed method against different hyperparameters in Section~\ref{sec:ablations}. Lastly, we provide a quantitative comparison of quality of posterior parameter samples in Section~\ref{sec:comp-post-qual}.   We refer the reader to Appendix for details regarding these experiments.

 \subsection{Baselines and Comparisons}
 \label{sec:baseline}

 We consider the state-of-the-art BPC methods using reverse-KL (BPC-rKL), forward-KL (BPC-fKL), and Wasserstein distance (BPC-W)~\citep{BPC,BPC_OG} for comparison. Further comparison with other coreset methods and dataset condensation is provided in the Appendix. All the baselines are implemented using the official codebase provided by respective methods if available, otherwise, we directly take the reported numbers. In cases, neither the codebase nor the numbers are reported, we exclude them from our tables.

 \subsection{Results and Comparison}
 \label{sec:performance_low_res}

\begin{table*}[!t]
\caption{\label{sota-comp-BPC}Comparison of the proposed method with BPC baselines. The results are noted in the form of (mean $\pm$ std. dev) where we have obtained test accuracy over five independent runs on the pseudo-coreset. The best performer across all methods is denoted in bold ($\boldsymbol{x\pm s}$). }
 \resizebox{\textwidth}{!}
 {
 \renewcommand{\arraystretch}{1.1}
 
\begin{tabular}{l|rr|rr|rr|rr|rr|rr}
\toprule
 &
  \textbf{ipc} &
  \textbf{Ratio(\%)} &
  \multicolumn{2}{|c} {\textbf{BPC-rKL(sghmc)}} &
  \multicolumn{2}{|c} {\textbf{BPC-W (sghmc)}} &
  \multicolumn{2}{|c} {\textbf{BPC-fKL (hmc)}} &
  \multicolumn{2}{|c} {\textbf{BPC-fKL (sghmc)}} &
  \multicolumn{2}{|c}{\textbf{Ours}} \\
 & 
  \multicolumn{1}{l} {}&
  \multicolumn{1}{l}{} &
  \multicolumn{1}{|c} {\textbf{Acc($\uparrow$)}} &
  \multicolumn{1}{c} {\textbf{NLL($\downarrow$)}} &
  \multicolumn{1}{|c} {\textbf{Acc($\uparrow$)}} &
  \multicolumn{1}{c} {\textbf{NLL($\downarrow$)}} &
  \multicolumn{1}{|c} {\textbf{Acc($\uparrow$)}} &
  \multicolumn{1}{c}{\textbf{NLL($\downarrow$)}} &
  \multicolumn{1}{|c}{\textbf{Acc($\uparrow$)}} &
  \multicolumn{1}{c}{\textbf{NLL($\downarrow$)}} &
  \multicolumn{1}{|c}{\textbf{Acc($\uparrow$)}} &
  \multicolumn{1}{c}{\textbf{NLL($\downarrow$)}} \\ \midrule
 &
  1 &
  0.017 &
  $74.80 \pm 1.17$ &
  $1.90 \pm 0.01$ &
  $83.59 \pm 1.49$ &
  $1.91 \pm 0.02$ &
  $90.46 \pm 1.50$ &
  $1.54\pm 0.03$ &
  $82.98 \pm 2.20$ &
  $1.87 \pm 0.03$ &
  \boldsymbol{$93.42 \pm 0.09$} &
  \boldsymbol{$1.53\pm 0.01$} \\
 &
  10 &
  0.17 &
  $95.27 \pm 0.17$ &
  $1.53 \pm 0.01$ &
  $91.72 \pm 0.55$ &
  $1.52 \pm 0.01$ &
  $89.80 \pm 0.82$ &
  $1.52 \pm 0.01$ &
  $92.05 \pm 0.42$ &
  \boldsymbol{$1.51 \pm 0.02$} &
  \boldsymbol{$97.71 \pm 0.24$} &
  $1.57 \pm 0.02$ \\
 \textbf{\multirow{-3}{*}{	\textbf{MNIST}}} &
  50 &
  0.83 &
  $94.18 \pm 0.26$ &
  $1.36 \pm 0.02$ &
  $93.72 \pm 0.55$ &
  $1.48 \pm 0.02$ &
  $95.58 \pm 1.63$ &
  $1.37 \pm 0.02$ &
  $93.63 \pm 1.80$ &
  $1.36 \pm 0.02$ &
  \boldsymbol{$98.91 \pm 0.22$} &
  \boldsymbol{$1.36 \pm 0.01$} \\ \midrule
 &
  1 &
 0.017 &
  $70.53 \pm 1.09$ &
  $2.47 \pm 0.02$ &
  $72.39 \pm 0.87$ &
  $2.10 \pm 0.01$ &
  \boldsymbol{$78.24 \pm 1.02$} &
  $1.95 \pm 0.04$ &
  $72.51 \pm 2.53$ &
  $2.30 \pm 0.02 $ &
  $77.29 \pm 0.50$ &
  \boldsymbol{$1.90\pm 0.03$} \\
 &
  10 &
  0.17 &
  $78.81 \pm 0.17$ &
  $1.64 \pm 0.01$ &
  $83.69 \pm 0.51$ &
  $1.64 \pm 0.03$ &
  $82.06 \pm 0.44$ &
  \boldsymbol{$1.53 \pm 0.02 $} &
  $83.29 \pm 0.55$ &
  $1.54 \pm 0.03 $ &
  \boldsymbol{$88.40 \pm 0.21$} &
  $1.56\pm0.01$ \\ 
\multirow{-3}{*}{\textbf{FMNIST}} &
  50 &
  0.83 &
  $76.97 \pm 0.59$ &
  $1.48 \pm 0.02$ &
  $74.41 \pm 0.48$ &
  $1.52 \pm 0.03$ &
  $82.40 \pm 0.35$ &
  $1.32 \pm 0.02$ &
  $74.82 \pm 0.52$ &
  $1.47 \pm 0.02 $ &
  \boldsymbol{$89.47 \pm 0.06$} &
  \boldsymbol{$1.30\pm0.02$} \\ \midrule
 &
  1 &
  0.014 &
  $18.34 \pm 1.79$ &
  $3.01 \pm 0.02$ &
  $33.52 \pm 1.15$ &
  $2.89 \pm 0.01$ &
  $48.02 \pm 5.62$ &
  $2.44 \pm 0.03$ &
  $21.48 \pm 6.58$ &
  $ 2.57 \pm 0.02 $ &
  \boldsymbol{$66.74 \pm 0.09$} &
  \boldsymbol{$2.38\pm0.04$} \\
 &
  10 &
  0.14 &
  $60.68 \pm 5.07$ &
  $2.00 \pm 0.01$ &
  $74.75 \pm 1.27$ &
  $1.95 \pm 0.02$ &
  $65.64 \pm 2.92$ &
  $2.13 \pm 0.01$ &
  $75.49 \pm 0.84$ &
  $1.84 \pm 0.01 $ &
  \boldsymbol{$82.32 \pm 0.56$} &
  \boldsymbol{$1.81\pm0.01$} \\
\multirow{-3}{*}{\textbf{SVHN}} &
  50 &
  0.7 &
  $78.27 \pm 0.62$ &
  $1.89 \pm 0.01$ &
  $79.49 \pm 0.54$ &
  $1.90 \pm 0.01$ &
  $79.60 \pm 0.53$ &
  $1.86 \pm 0.01$ &
  $77.08 \pm 1.80$ &
  \boldsymbol{$1.72 \pm 0.01 $ }&
  \boldsymbol{$88.41 \pm 0.12$} &
  $1.88\pm0.02$ \\ \midrule
 &
  1 &
  0.02 &
  $21.62 \pm 0.83$ &
  $2.57 \pm 0.01$ &
  $29.34 \pm 1.21$ &
  $2.14 \pm 0.03$ &
  $35.57 \pm 0.95$ &
  $1.97 \pm 0.03$ &
  $29.30 \pm 1.10$ &
  $ 2.10 \pm 0.03 $ &
  \boldsymbol{$46.87 \pm 0.20$} &
  \boldsymbol{$1.87\pm0.02$} \\
 &
  10 &
  0.2 &
  $37.89 \pm 1.54$ &
  $2.13 \pm 0.02$ &
  $48.90 \pm 1.72$ &
  $1.73 \pm 0.02$ &
  $43.07 \pm 1.06$ &
  $1.89\pm 0.02$ &
  $49.85 \pm 1.37$ &
  $ 1.73 \pm 0.01$ &
  \boldsymbol{$56.39 \pm 0.70$} &
  \boldsymbol{$1.72\pm0.03$} \\
\multirow{-3}{*}{\textbf{Cifar10}} &
  50 &
  1 &
  $37.54 \pm 1.32$ &
  $ 1.93 \pm 0.03$ &
  $46.17 \pm 0.67$ &
  $1.62 \pm 0.02$ &
  $50.92 \pm 1.49$ &
  $1.70\pm0.03$ &
  $42.30 \pm 2.87$ &
  \boldsymbol{$ 1.54 \pm 0.01 $} &
  \boldsymbol{$71.93 \pm 0.17$} &
  $1.57\pm 0.03$ \\ \midrule
 &
  1 &
  0.2 &
  $3.56 \pm 0.04$ &
  $4.69 \pm 0.02$ &
  $12.19 \pm 0.22$ &
  $4.20 \pm 0.01 $ &
  $7.57 \pm 0.54$ &
  $4.25\pm0.04$ &
  $12.07 \pm 0.16$ &
  $4.27 \pm 0.02$ &
  \boldsymbol{$23.97 \pm 0.11$} &
  \boldsymbol{$4.01\pm 0.02$} \\
\multirow{-2}{*}{\textbf{Cifar100}} &
  10 &
  2 &
  - &
   &
  - &
   &
  - &
   &
  - &
  - &
  \boldsymbol{$28.42 \pm 0.24$} &
  \boldsymbol{$3.14\pm 0.02$ }\\ \midrule
 &
  1 &
  0.2 &
  - &
   &
  - &
   &
  - &
   &
  - &
  - &
  \boldsymbol{$8.39 \pm 0.07$} &
  \boldsymbol{$4.72\pm 0.01$} \\
\multirow{-2}{*}{\textbf{T-ImageNet}} &
  10 &
  2 &
  - &
   &
  - &
   &
  - &
   &
  - &
  - &
  \boldsymbol{$17.82 \pm 0.39$} &
  \boldsymbol{$3.64\pm0.05$} \\
\bottomrule
\end{tabular}
}
\end{table*}

\begin{figure*}[!t]
\centering
   \begin{subfigure}[t]{0.239\textwidth}
   \begin{minipage}{\textwidth}
   \includegraphics[keepaspectratio, width=\textwidth]{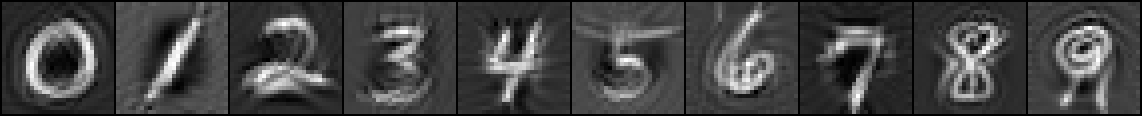}
   \caption{Images generated with ipc=1 for MNIST}
   \includegraphics[keepaspectratio, width=\textwidth]{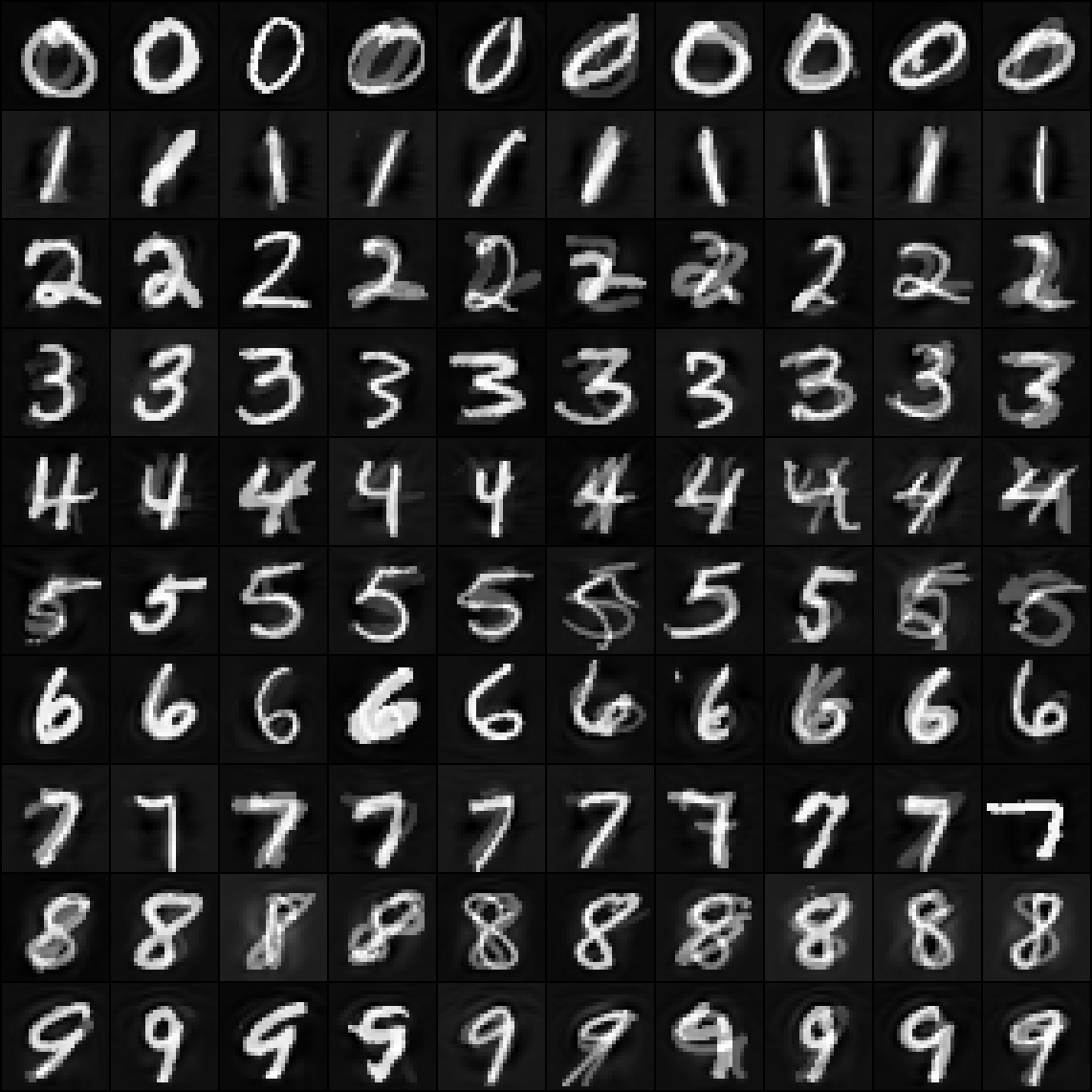}
       \caption{Images generated with ipc=10 for MNIST}
   
    \end{minipage}
\end{subfigure}
~
\begin{subfigure}[!t]{0.239\textwidth}
\begin{minipage}{\textwidth}
    
       \includegraphics[keepaspectratio, width=\textwidth]{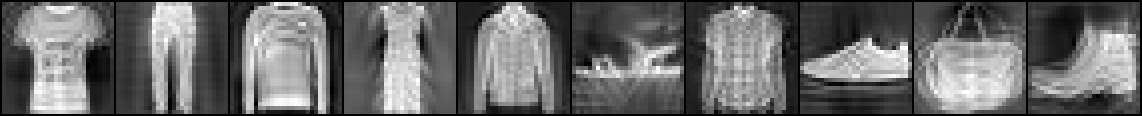}
       \caption{Images generated with ipc=1 for FMNIST}
   \includegraphics[keepaspectratio, width=\textwidth]{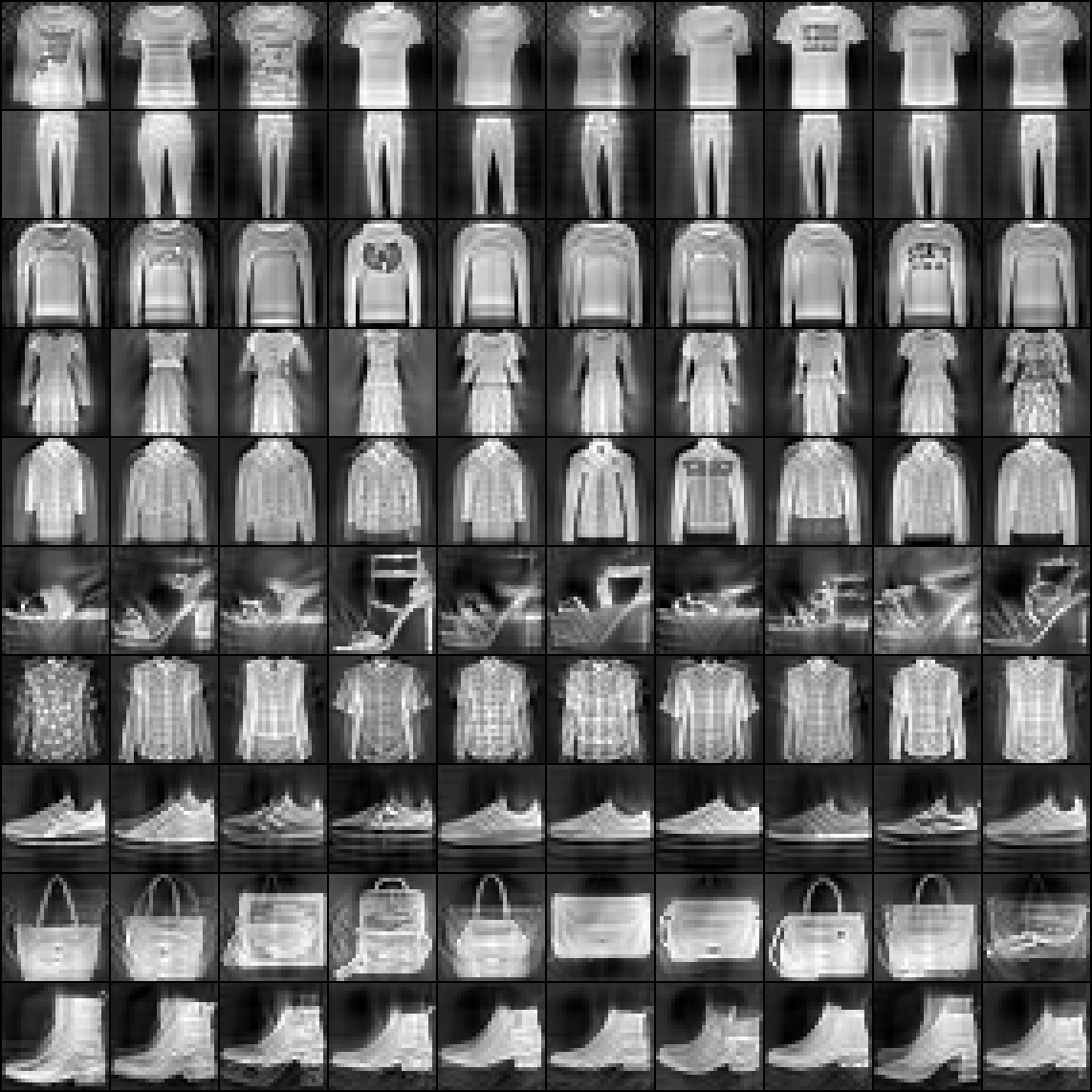}
   \caption{Images generated with ipc=10 for FMNIST}

\end{minipage}
\end{subfigure}
~
\begin{subfigure}[t]{0.239\textwidth}
\begin{minipage}{\textwidth}
   
   \includegraphics[keepaspectratio, width=\textwidth]{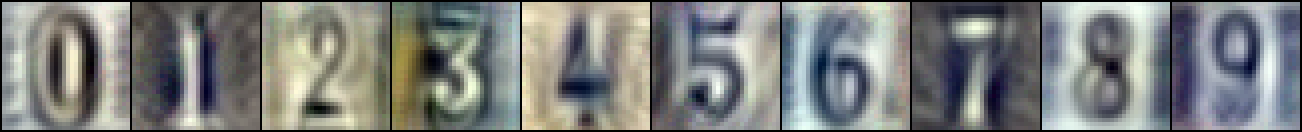}
   \caption{Images generated with ipc=1 for SVHN}
   \includegraphics[keepaspectratio, width=\textwidth]{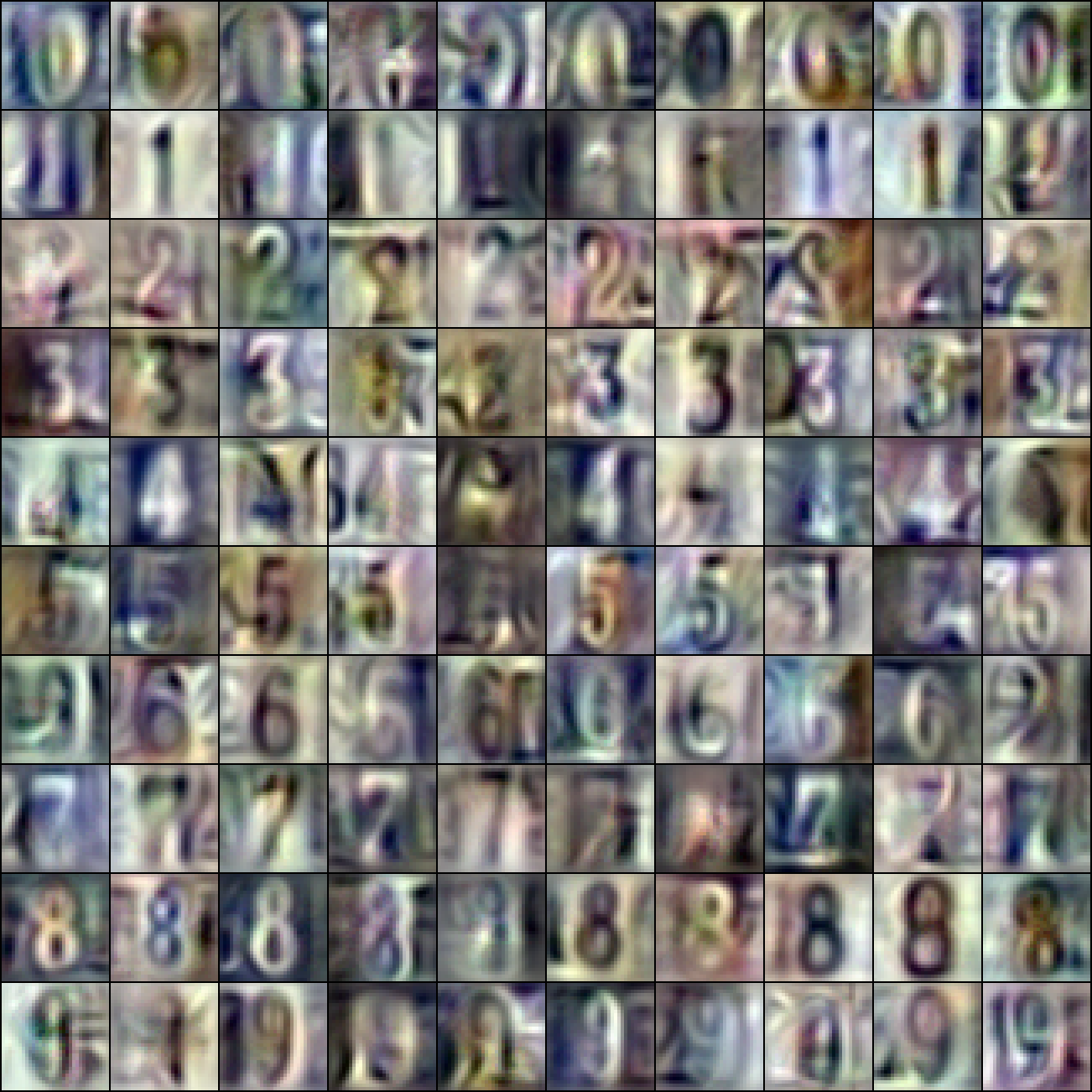}
   \caption{Images generated with ipc=10 for SVHN}
\end{minipage}
\end{subfigure}
~
\begin{subfigure}[t]{0.239\textwidth}
\begin{minipage}{\textwidth}

      \includegraphics[keepaspectratio, width=\textwidth]{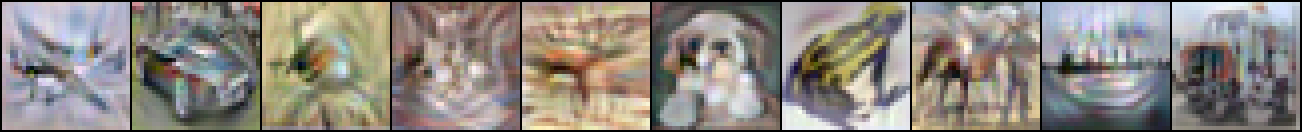}
      \caption{Images generated with ipc=1 for CIFAR10}
   \includegraphics[keepaspectratio, width=\textwidth]{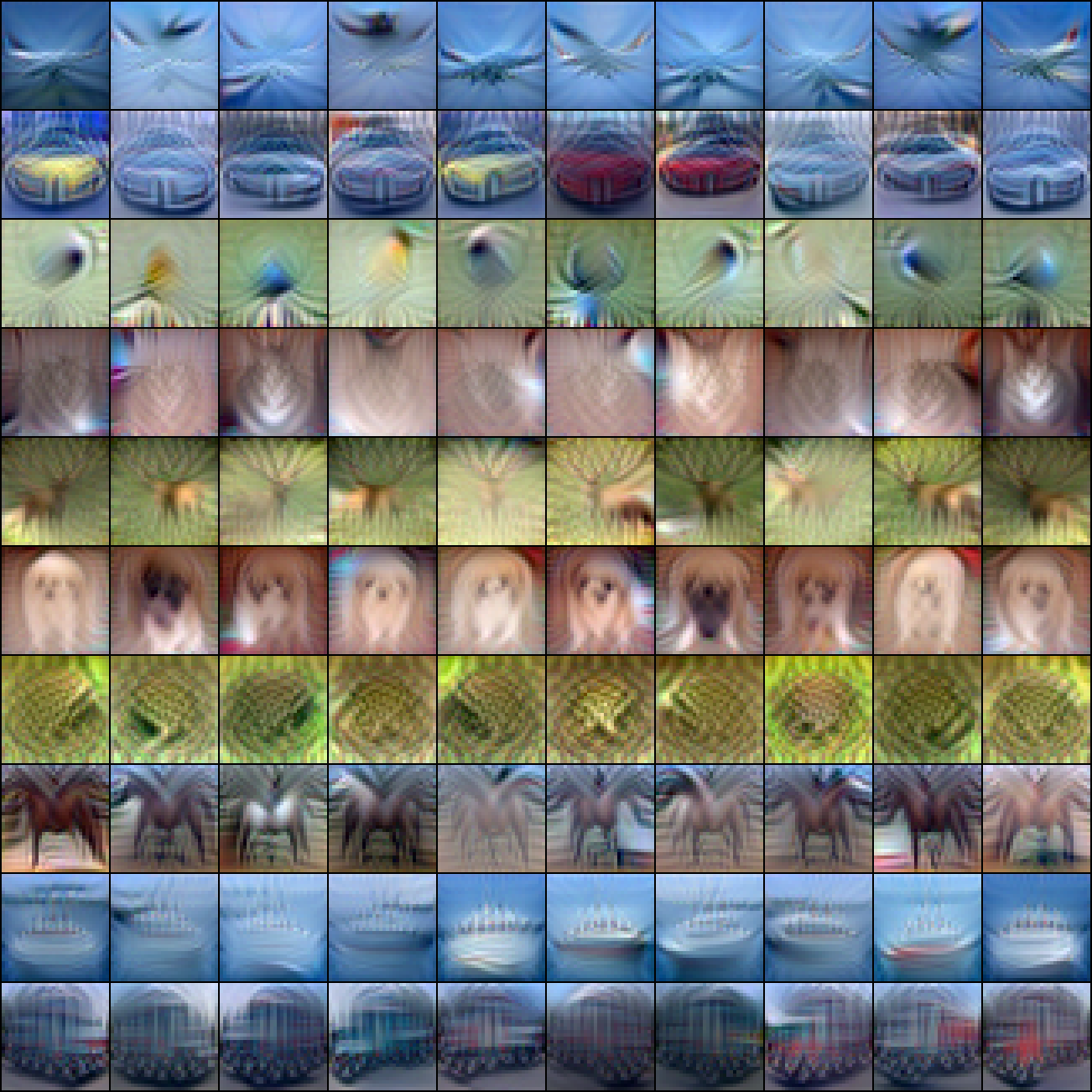}
    \caption{Images generated with ipc=10 for CIFAR10}
\end{minipage}
\end{subfigure}
\caption{\label{viz-ipc-10}Visualizations of pseudo-coreset generated from our method with one image per class (top) and ten images per class (bottom) for MNIST, FMNIST, SVHN and CIFAR10. It can be seen that the class labels are identifiable to a large extent.}
\end{figure*}

Table~\ref{sota-comp-BPC} presents the results of the comparative analysis between our approach and other BPC baselines.
We observe that the proposed method significantly outperforms all the BPC baselines by large margins. For instance, we observe an improvement of $11.3\%$, $6.54\%$, and $21.01\%$ in accuracy for CIFAR10 with ipc values of 1, 10, and 50, respectively. Additionally, there is a decrease of 0.1 and 0.01 points in negative log-likelihood for ipc values of 1 and 10, respectively, in comparison to the best-performing BPC baseline.  Similarly, on SVHN, we notice an improvement in accuracy and negative log-likelihood. Specifically, we observe gains of $18.72\%$, $6.83\%$ in accuracy and reduction of $0.06$, $0.03$ point in negative log-likelihood for ipc 1, 10 respectively, compared to the BPC counterparts. A similar trend can be seen for MNIST and FMNIST as well~\footnote{We observe that the proposed method has smaller variance compared to other methods. Upon analysis, we attribute this behavior to the rapid decrease in the gradient of the energy function ($(\nabla_{\theta}E(\theta,\tilde{\rvx})$) with respect to parameters in langevin dynamics. This indicates that the sampled parameter stays near the initial parameter value from the expert trajectory, which is based on confident predictions derived from SGD trajectories trained on the full dataset. Therefore, the reduced variance observed is likely a result of this confidence.}. 
We attribute this boost in performance to the flexible formulation of the proposed method.

We present the qualitative visualizations for MNIST, FMNIST, SVHN, and CIFAR10 datasets with 1 image per class and 10 image per class in Fig.~\ref{viz-ipc-10}. It can be seen that the constructed pseudo-coreset is identifiable but inherits some artifacts due to the constraints on the dataset size. As the number of images per class increases, the model can induce more variations across all the classes and thus produce a diverse pseudo-coreset. Additional qualitative visualizations for pseudo-coreset generated with 50 images per class on CIFAR100 and T-ImageNet dataset are presented in the Appendix.

\subsection{Results on Out of Distribution (OOD) dataset}
\label{sec:ood-results}
We present the results of the proposed method on out-of-distribution (OOD) dataset in Table~\ref{tab:ood-performance}.
We use CIFAR10-C~\citep{CIFAR10-C} dataset for this experiment.
In particular, we sample the parameters from the pseudo-coreset posterior obtained using clean CIFAR10 (ipc=10) and perform inference on the corrupted CIFAR10-C, which consists of CIFAR10 images afflicted with different types of corruption including Gaussian Blur, Gaussian Noise, etc.
It is evident from Table~\ref{tab:ood-performance} that our method demonstrates robustness to various types of corruption and exhibits superior performance compared to other baselines. Notably, for corruptions like Gaussian Blur, our method achieves a $1.63\%$ increase in accuracy and a $0.17$-point reduction in negative log-likelihood compared to the best-performing BPC baseline. Likewise, for JPEG Compression, Zoom Blur, and Defocus Blur, our method yields an improvement of $0.07\%$, $2.08\%$, and $0.43\%$ in accuracy, along with a reduction of $0.14$, $0.05$, and $0.06$ points in negative log-likelihood, respectively. The robustness of the proposed method to different forms of corruption highlights its ability to provide a better approximation of underlying posterior distribution when compared to other baselines. \\
We further test all the BPC methods against $\ell_{\infty}$ adversarial attack~\citep{robustbench}. We report the clean accuracy and robust accuracy respectively for each method. Our observations could be found in Table~\ref{tab:adv-attack}. We see that under $\ell_{\infty}$ attack, the performance of all BPC methods drop significantly. This is perhaps due to the fact that these methods don't explicitly take robustness into account while constructing pseudo-coresets. However, we observe even under performance drop, our method gives best robust accuracy as compared to other BPC methods.

\begin{table*}[!t]
\centering
\caption{Comparison of the proposed method with BPC baselines for the performance on  out-of-distribution data. The classifier model is trained on pseudo-coresets generated using CIFAR10. However, the model is evaluated on CIFAR10-C dataset with different types of corruption.}
\resizebox{0.95\textwidth}{!}{%
\begin{tabular}{l|rr|rr|rr|rr|rr}
\toprule
\multirow{2}{*}{\textbf{Corruption}} & \multicolumn{2}{c|}{\textbf{BPC-rKL (sghmc)}} & \multicolumn{2}{c|}{\textbf{BPC-W (sghmc)}} & \multicolumn{2}{c|}{\textbf{BPC-fKL (hmc)}} & \multicolumn{2}{c|}{\textbf{BPC-fKL (sghmc)}} & \multicolumn{2}{c}{\textbf{Ours}}                                    \\ 
                            & \textbf{Acc($\uparrow$)}               & \textbf{NLL($\downarrow$)}              & \textbf{Acc($\uparrow$)}     & \textbf{NLL($\downarrow$)}    & \textbf{Acc($\uparrow$)}     & \textbf{NLL($\downarrow$)}    & \textbf{Acc($\uparrow$)}      & \textbf{NLL($\downarrow$)}     & \textbf{Acc($\uparrow$)}                  & \textbf{NLL($\downarrow$)}                 \\ \midrule
Gaussian Blur               & $31.02\pm2.65$        & $2.13\pm0.77$       & $35.66\pm1.21$       & $2.04\pm0.12$       & $34.76\pm1.86$       & $1.89\pm0.04$    & $39.73\pm2.72$        & $1.94\pm0.05$       & \boldsymbol{$41.36\pm 0.72$} & \boldsymbol{$1.73 \pm 0.83$} \\
Gaussian Noise              & $25.49\pm1.89 $       &$ 2.28\pm0.08$        & $33.21\pm0.89$       & $2.11\pm0.03$       & $36.70\pm1.01$         & $1.86\pm0.02$       & $35.71\pm2.29$        & $2.00\pm0.05$        & \boldsymbol{$38.01\pm 1.26$} & \boldsymbol{$1.82 \pm 0.11$} \\
JPEG Compression            & $30.40\pm0.90 $        &$ 2.13\pm0.02$        & $26.33\pm1.34$       & $2.26\pm0.04$       & $36.20\pm1.92$        & $1.85\pm0.03$       & $37.26\pm2.87$        & $1.95\pm0.06$        & \boldsymbol{$37.33\pm 0.19$} & \boldsymbol{$1.71 \pm 0.03$} \\
Snow                        & $26.85\pm1.71$        &$ 2.20\pm0.07$         & $37.50\pm3.50$         & $1.93\pm0.08$       & $33.99\pm1.91$       & \boldsymbol{$1.91\pm0.03$}       & $35.68\pm2.71$        & $2.00\pm0.07$         & \boldsymbol{$37.84\pm 0.64$} & $1.91 \pm 0.05$ \\
Impulsive Noise             & $28.39\pm1.48$        &$ 2.15\pm0.06$        & $36.71\pm1.93$       & $1.96\pm0.05$       & $33.81\pm1.58$       & $1.94\pm0.02$       & \boldsymbol{$38.26\pm2.34$}        & $1.92\pm0.05$        & $37.98\pm 2.15$ & \boldsymbol{$1.89 \pm 0.07$} \\
Zoom Blur                   & $31.74\pm1.24$        &$ 2.09\pm0.04$        & $36.22\pm2.08$       & $1.99\pm0.05$       & $31.30\pm3.64$        & $1.98\pm0.08$       & $35.05\pm2.90$        & $2.04\pm0.07$        & \boldsymbol{$38.30\pm 0.77$} & \boldsymbol{$1.93 \pm 0.13$} \\
Pixelate                    & $28.98\pm2.26$        &$ 2.19\pm0.07$        & $27.98\pm1.77$       & $2.20\pm0.05$       & $35.59\pm1.94$       & \boldsymbol{$1.88\pm0.03$}       & \boldsymbol{$39.14\pm3.15$}        & $1.93\pm0.06$        & $38.97\pm 1.51$ & $1.92 \pm 0.07$ \\
Speckle Noise               & $29.88\pm0.59$        &$ 2.09\pm0.02$        & $33.33\pm2.18$       & $2.05\pm0.05$       & $34.37\pm2.02$       & $1.90\pm3.57$       & $40.54\pm1.93$        & \boldsymbol{$1.89\pm0.04$}        & \boldsymbol{$42.66\pm 0.83$} & $1.95 \pm 0.03$ \\
Defocus Blur                & $27.57\pm1.31$        &$ 2.20\pm0.05$        & $33.80\pm4.21$       & $2.09\pm0.11$       & $33.60\pm2.93$        & $1.93\pm0.06$       & $36.72\pm3.68$        & $1.99\pm0.08$        & \boldsymbol{$37.15\pm 1.03$} & \boldsymbol{$1.87 \pm 0.04$} \\
Motion Blur                 & $17.38\pm2.51$        &$ 2.73\pm0.14$        & $35.22\pm3.35$       & $2.01\pm0.08$       & $34.33\pm1.89$       & $1.92\pm0.04$      & $35.24\pm3.30$        & $2.01\pm0.05$        & \boldsymbol{$37.06\pm 0.49$} & \boldsymbol{$1.92 \pm 0.04$} \\ \bottomrule
\end{tabular}%
}
\label{tab:ood-performance}
\end{table*}

\begin{table*}[!t]
\centering
\caption{Comparison of the proposed method with BPC baselines for the performance against $\ell_{\infty}$ attack on CIFAR10 with 10 ipc.}
\resizebox{0.9\textwidth}{!}{%

\begin{tabular}{rr|rr|rr|rr}
\toprule
                            \multicolumn{2}{c|}{\textbf{BPC-rkl}}    & \multicolumn{2}{c|}{\textbf{BPC-W}}      & \multicolumn{2}{c|}{\textbf{BPC-fkl}}    & \multicolumn{2}{c}{\textbf{Ours}}                   \\ 
                            \textbf{Clean Acc} & \textbf{Robust Acc} & \textbf{Clean Acc} & \textbf{Robust Acc} & \textbf{Clean Acc} & \textbf{Robust Acc} & \textbf{Clean Acc}       & \textbf{Robust Acc}       \\ \midrule
 $37.89 \pm 1.54$   & $13.54 \pm 1.21$    & $48.90 \pm 1.72$    & $15.62 \pm 0.92$    & $49.85 \pm 1.37$   & $19.30 \pm 1.39$    & $\mathbf{56.39 \pm 0.70}$ & $\mathbf{22.81 \pm 1.28}$ \\ \bottomrule
\end{tabular}%
\label{tab:adv-attack}
}
\end{table*}

\begin{table*}[!t]
  \begin{minipage}{0.45\linewidth}
    \centering
    \caption{\label{cross-architecture}Cross-architecture generalization analysis of BPC methods.}

\renewcommand{\arraystretch}{1.3}
\resizebox{\textwidth}{!}{
\begin{tabular}{l|rrrr}
\toprule
 & \textbf{ConvNet} & \textbf{ResNet} & \textbf{VGG} & \textbf{AlexNet} \\ \midrule
    \textbf{Ours} &
  \boldsymbol{$56.39 \pm 0.70$} &
  \boldsymbol{$41.65 \pm 1.03$} &
  \boldsymbol{$47.51 \pm 0.89$} &
  \boldsymbol{$30.58 \pm 1.43$} \\
\textbf{BPC-fKL(hmc)} &
  $44.34 \pm 1.11$ &
  $10.15 \pm 0.21$ &
  $10.43 \pm 0.33$ &
  $12.21 \pm 0.18$ \\
\textbf{BPC-rKL(sghmc)} &
  $34.48 \pm 0.48$ &
  $10.06 \pm 0.08$ &
  $10.26 \pm 0.35$ &
  $11.02 \pm 0.12$ \\ \bottomrule
\end{tabular}}
  \end{minipage}%
  \hspace{0.3cm}
  \begin{minipage}{0.58\linewidth}
    \centering
    \caption{Performance comparison of the proposed method and other BPC baselines for different parameterized architectures.}
\centering
\resizebox{\textwidth}{!}{
\begin{tabular}{l|rrrrrrr}
\toprule
\textbf{Methods} &  &  \textbf{CN-D3W128} & \textbf{CN-D3W256} & \textbf{CN-D5W128} &  \textbf{AlexNet} &  \textbf{VGG11} &  \textbf{ResNet} \\
 &  &  320,010 &1,229,834 &  596,490 &  1,872,202 &  9,231,114 & 11,173,962 \\
 \midrule
 \textbf{Ours} &  & \boldsymbol{$56.39 \pm 0.70$} & \boldsymbol{$55.93\pm 1.30$} & \boldsymbol{$56.01 \pm 0.69$} & \boldsymbol{$52.88 \pm 1.39$} & \boldsymbol{$49.26 \pm 2.33$} & \boldsymbol{$48.67 \pm 0.52$} \\
 \textbf{BPC-rKL (sghmc)} &  & $37.89 \pm 1.54$ & $35.82 \pm 1.88$ & $35.92 \pm 1.88$ & $32.60 \pm 1.45$ &  $27.66 \pm 0.73$ & $24.98 \pm 1.53$ \\
 \textbf{BPC-W (sghmc)} &  & $48.90 \pm 1.72$ & $43.71 \pm 1.42$ & $46.01\pm 0.92$ & $39.01 \pm 0.51$ & $35.11 \pm 1.82$ & $32.84 \pm 1.38$ \\
\textbf{BPC-fKL (hmc)} &  & $49.85 \pm 1.37$ & $45.87 \pm 0.78$ & $47.92 \pm 1.27$ & $41.22 \pm 1.62$ & $37.05 \pm 1.24$ & $35.10 \pm 2.03$ \\
 \bottomrule

\end{tabular}}
\label{tab:diff-arch}
  \end{minipage}
\end{table*}

\subsection{Results on Cross-Architecture Experiments}
\label{sec:cross-arch}
Here, we present the cross-architecture results pertaining to various BPC methods. In these experiments, we construct the pseudo-coreset using the said ConvNet model, while during inference, we use different architectures such as ResNet~\citep{resnet}, VGG-Net~\citep{vgg} and AlexNet~\citep{alexnet} for evaluation. We perform these experiments for CIFAR10 (ipc = 10).
The results of the cross-architecture experiments are presented in Table~\ref{cross-architecture}. It can be seen that previous BPC methods fail to generalize across different network architectures, whereas our method demonstrates the ability to adapt to various architectures. For instance, the performance of BPC-fKL and BPC-rKL drop by $34.19\%$ and $24.42\%$ respectively on ResNet, resulting in random predictions with an accuracy of almost $10\%$, whereas our method observes a drop of only $14.74\%$ while giving an accuracy of $41.65\%$. 

\subsection{Effect of Different number of Parameters}
\label{sec:diff-arch}
Lastly, we analyze the performance of BPC methods across differently parameterized networks. Specifically, we generate pseudo-coresets for CIFAR10 (ipc=10) by employing ConvNets with different parameter configurations. These configurations encompass ConvNets with different depth and width. We also conduct a comparative analysis with other deep learning architectures, including AlexNet~\citep{alexnet}, VGG11~\citep{vgg}, and ResNet~\citep{resnet}. Bayesian inference techniques generally encounter scalability issues when dealing with large parametric networks~\citep{handsOnBayesian}. This experiment is conducted to ascertain the impact of both large and small architectures on the performance of pseudo-coresets.      

The results for different parameterized architectures are presented in Table~\ref{tab:diff-arch}. Here, CN-DxWy denotes a ConvNet architecture with a depth of `x' and width of `y'. It is evident from the results that the performance of all BPC methods declines as the number of parameters in the architectures increases. However, our model exhibits relatively better performance in comparison to other methods. Specifically, while our method demonstrates a $7.72\%$ decrease in performance for the ResNet architecture, other BPC baselines such as BPC-fKL, BPC-W, and BPC-rKL experience declines of approximately $14.75\%$, $16.06\%$, and $12.91\%$, respectively. This observation underscores the greater tolerance of our method to large parametric models when compared to other baselines. We again highlight that the performance gain achieved by our proposed method can be attributed to the current formulation, which can generate better approximation to the true posterior.

\subsection{Ablation Study}
\label{sec:ablations}
Next, we provide an empirical analysis of the effect of different hyperparameters on the proposed method. Particularly, we observe that there are two key hyperparameters that affect the performance of the proposed method: the covariance matrix $\Sigma_{\rvx}$ (see Eq.~\ref{eq:var_approx})   and the number of MCMC steps $k$ (see Eq.~\ref{eq:final_cd}). Hence, we ablate our method against these two important hyperparameters and report the results in Table~\ref{tab:main-ablation}.

Further, we verify our claim that the proposed method provides a good trade-off between performance and computational cost. Particularly, as seen in Eq.~\ref{eq:fkl-loss}, one can resort to MCMC methods to evaluate the second term of the said expression, however, since this can be computationally expensive in the high-dimensional parameter space, we resort to using contrastive-divergence that allows one to use finite step MCMC steps making the proposed method computationally less expensive. To verify this, we compare our method against pseudo-coresets obtained using Eq.~\ref{eq:fkl-loss} alongwith MCMC methods. Particularly, we employ HMC~\citep{sghmc} and Kronecker-Factorised Laplace (KFL)~\citep{kfl} for this purpose. Our observations are noted in Table~\ref{tab:mcmc-comp}.We compare GPU Memory usage, Iteration time and Accuracy for each of these method against ours. We see that both HMC and KFL consume more memory and time compared to our method. However, HMC provides marginally better result and KFL gives slightly worse result compared to our method. This verifies our claim.

\begin{table}[!t]
    \centering
    \caption{Ablation study of the proposed method against different hyperparameters on CIFAR10 with 10 ipc.}
    \resizebox{\columnwidth}{!}{
    \begin{tabular}{rrrrrr}
        \toprule
        \multicolumn{6}{c}{\textbf{Ablation on $\Sigma_{\rvx}$}}                                                                                                                                                                                     \\ \midrule
        \multicolumn{2}{c|}{\boldsymbol{$\Sigma_{\rvx}^{1/2} = 0.01I$}}                     & \multicolumn{2}{c|}{\boldsymbol{$\Sigma_{\rvx}^{1/2} = 0.001I$}}                    & \multicolumn{2}{c}{\boldsymbol{$\Sigma_{\rvx}^{1/2} = 0.0001I$}}                   \\ 
        \multicolumn{1}{c}{\textbf{Acc}}     & \multicolumn{1}{c|}{\textbf{NLL}}    & \multicolumn{1}{c}{\textbf{Acc}}     & \multicolumn{1}{c|}{\textbf{NLL}}    & \multicolumn{1}{c}{\textbf{Acc}}     & \multicolumn{1}{c}{\textbf{NLL}}    \\ \midrule
        \multicolumn{1}{c}{$50.18 \pm 0.50$} & \multicolumn{1}{c|}{$1.94 \pm 0.05$} & \multicolumn{1}{c}{$\mathbf{56.39 \pm 0.70}$} & \multicolumn{1}{c|}{$\mathbf{1.72 \pm 0.03}$} & \multicolumn{1}{c}{$54.18 \pm 0.23$} & \multicolumn{1}{c}{$1.86 \pm 0.02$} \\ \bottomrule
        \multicolumn{6}{c}{}                                                                                                                                                                                                                        \\ \toprule
        \multicolumn{6}{c}{\textbf{Ablation on $k$ (or $L$ in Algorithm 1)}}                                                                                                                                                                      \\ \midrule
        \multicolumn{2}{c|}{\boldsymbol{$k = 10$}}                                       & \multicolumn{2}{c|}{\boldsymbol{$k = 50$}}                                       & \multicolumn{2}{c}{\boldsymbol{$k = 100$}}                                      \\ 
        \multicolumn{1}{c}{\textbf{Acc}}     & \multicolumn{1}{c|}{\textbf{NLL}}    & \multicolumn{1}{c}{\textbf{Acc}}     & \multicolumn{1}{c|}{\textbf{NLL}}    & \multicolumn{1}{c}{\textbf{Acc}}     & \multicolumn{1}{c}{\textbf{NLL}}    \\ \midrule
        \multicolumn{1}{c}{$43.77 \pm 0.98$} & \multicolumn{1}{c|}{$1.78 \pm 0.02$} & \multicolumn{1}{c}{$51.27 \pm 1.01$} & \multicolumn{1}{c|}{$1.78 \pm 0.02$} & \multicolumn{1}{c}{$\mathbf{56.39 \pm 0.70}$} & \multicolumn{1}{c}{$\mathbf{1.72 \pm 0.03}$} \\ \bottomrule
        \end{tabular}
        }
    \label{tab:main-ablation}
\end{table}

\begin{table}[!t]
    \caption{Efficiency and Accuracy comparison of the proposed method against pseudo-coresets obtained via direct MCMC estimation of second term of Eq.~\ref{eq:fkl-loss} on CIFAR10 with 10 ipc.}
    \centering
    \resizebox{\columnwidth}{!}{
    \begin{tabular}{ccc|ccc|ccc}
        \toprule
         \multicolumn{3}{c|}{\textbf{HMC}~\citep{sghmc}}                           & \multicolumn{3}{c|}{\textbf{KFL}~\citep{kfl}}  & \multicolumn{3}{c}{\textbf{Ours}}                          \\
                                    \textbf{GPU (GB)} & \textbf{Time (s)} & \textbf{Acc} & \textbf{GPU (GB)} & \textbf{Time (s)} & \textbf{Acc} & \textbf{GPU (GB)} & \textbf{Time (s)} & \textbf{Acc} \\ \midrule 
         52.67                    & 13.82             & $57.09$      & 61.05                    & 20.17             & $53.98$      & 38.28                    & 0.75              & $56.39$      \\ \bottomrule
        \end{tabular}%
    }
    \label{tab:mcmc-comp}
\end{table}

\subsection{Comparison of Posterior Quality}
\label{sec:comp-post-qual}
Lastly, we provide a quantitative comparison for the quality of posteriors obtained using our method and other baselines to substantiate our claims. Note that the true parameter posteriors are intractable and are generally unknown for complex deep networks. Hence, it is difficult to make comparisons against such gold standard posterior for complex networks. However, to provide a comprehensive understanding of the quality of our obtained posteriors, we report the Expected Calibration Error (ECE)~\citep{ECE} and Brier score~\citep{brier}. These metrics, akin to those presented in Table~6 of \citet{BPC}, serve as well-established benchmarks for evaluating posterior quality. These results are listed in Table~\ref{tab:callib-results}. We see that the proposed method performs the best amongst all the baselines, further ensuring that the proposed method is effective in the qualitative sense as well. 

\begin{table}[!t]
    \centering
    \caption{Comparison of ECE ($\downarrow$) and Brier Score ($\downarrow$) for the proposed method against other baselines}
    \resizebox{\columnwidth}{!}{
    \begin{tabular}{rr|rr|rr|rr}
        \toprule
                          \multicolumn{2}{c|}{\textbf{BPC-rkl}}     & \multicolumn{2}{c|}{\textbf{BPC-W}}       & \multicolumn{2}{c|}{\textbf{BPC-fkl}}     & \multicolumn{2}{c}{\textbf{Ours}}                        \\ 
                          \multicolumn{1}{c}{\textbf{ECE}}       & \multicolumn{1}{c|}{\textbf{Brier Score}} & \multicolumn{1}{c}{\textbf{ECE}}       & \multicolumn{1}{c|}{\textbf{Brier Score}} & \multicolumn{1}{c}{\textbf{ECE}}       & \multicolumn{1}{c|}{\textbf{Brier Score}} & \multicolumn{1}{c}{\textbf{ECE}}                & \multicolumn{1}{c}{\textbf{Brier Score}}        \\ \midrule
         $0.1183\pm 0.0038$ & $0.7988\pm 0.0038$   & $0.1457\pm 0.0110$ & $0.8030\pm 0.0049$   & $0.1538\pm 0.0049$ & $0.7231\pm 0.0049$   & $\mathbf{0.1092\pm 0.0052}$ & $\mathbf{0.6755\pm 0.0042}$ \\ \bottomrule
    \end{tabular}
    }
    \label{tab:callib-results}
\end{table}

\section{Conclusion}

In this work, we propose a novel approach to generate pseudo-coreset using contrastive divergence. Our approach addresses the need to approximate the posterior of pseudo-coreset and uses a finite number of steps in MCMC methods to sample the parameters from the underlying posterior distribution.
Subsequently, these parameters are used to construct pseudo-coreset via contrastive divergence. The empirical evidence presented in our study illustrates that our proposed method surpasses previous BPC baselines by substantial margins across multiple datasets.

\textbf{\textit{Limitations and Future Work}:}
While our approach effectively removes variational assumptions associated with the pseudo-coreset posterior and utilizes MCMC methods for parameter sampling, our study still relies on certain assumptions about the posterior of the original dataset. Since there remains a significant performance gap between the pseudo-coreset and the original dataset, a potential avenue for future research could be to relax these assumptions to enhance the performance of BPC methods.

\textbf{\textit{Broader Impact}:}
BPC methods have positive applications in democratization and privacy-related concerns by reducing the dependence on the original dataset. 
We don't believe that our method has any associated negative societal impact.

\begin{acknowledgements}
    This work was supported (in part for setting up the GPU compute) by the Indian Institute of Science through a start-up grant. Prathosh is supported by Infosys Foundation Young investigator award. Piyush is supported by Government of India via Prime minister’s research fellowship.
\end{acknowledgements}








\balance
\bibliography{uai2024-template}

\begin{thebibliography}{104}
\providecommand{\natexlab}[1]{#1}
\providecommand{\url}[1]{\texttt{#1}}
\expandafter\ifx\csname urlstyle\endcsname\relax
  \providecommand{\doi}[1]{doi: #1}\else
  \providecommand{\doi}{doi: \begingroup \urlstyle{rm}\Url}\fi

\bibitem[Agarwal et~al.(2005)Agarwal, Har-Peled, Varadarajan, et~al.]{agarwal2005geometric}
Pankaj~K Agarwal, Sariel Har-Peled, Kasturi~R Varadarajan, et~al.
\newblock Geometric approximation via coresets.
\newblock \emph{Combinatorial and computational geometry}, 52\penalty0 (1):\penalty0 1--30, 2005.

\bibitem[Ahn et~al.(2012)Ahn, Korattikara, and Welling]{ahn2012bayesian}
Sungjin Ahn, Anoop Korattikara, and Max Welling.
\newblock Bayesian posterior sampling via stochastic gradient fisher scoring.
\newblock \emph{arXiv preprint arXiv:1206.6380}, 2012.

\bibitem[Amodei et~al.(2016)Amodei, Ananthanarayanan, Anubhai, Bai, Battenberg, Case, Casper, Catanzaro, Cheng, Chen, et~al.]{amodei2016deep}
Dario Amodei, Sundaram Ananthanarayanan, Rishita Anubhai, Jingliang Bai, Eric Battenberg, Carl Case, Jared Casper, Bryan Catanzaro, Qiang Cheng, Guoliang Chen, et~al.
\newblock Deep speech 2: End-to-end speech recognition in english and mandarin.
\newblock In \emph{International conference on machine learning}, pages 173--182. PMLR, 2016.

\bibitem[Bachem et~al.(2015)Bachem, Lucic, and Krause]{bachem2015coresets}
Olivier Bachem, Mario Lucic, and Andreas Krause.
\newblock Coresets for nonparametric estimation-the case of dp-means.
\newblock In \emph{International Conference on Machine Learning}, pages 209--217. PMLR, 2015.

\bibitem[Bachem et~al.(2016)Bachem, Lucic, Hassani, and Krause]{bachem2016approximate}
Olivier Bachem, Mario Lucic, S~Hamed Hassani, and Andreas Krause.
\newblock Approximate k-means++ in sublinear time.
\newblock In \emph{Proceedings of the AAAI conference on artificial intelligence}, volume~30, 2016.

\bibitem[Bachem et~al.(2018)Bachem, Lucic, and Krause]{bachem2018scalable}
Olivier Bachem, Mario Lucic, and Andreas Krause.
\newblock Scalable k-means clustering via lightweight coresets.
\newblock In \emph{Proceedings of the 24th ACM SIGKDD International Conference on Knowledge Discovery \& Data Mining}, pages 1119--1127, 2018.

\bibitem[Bardenet et~al.(2014)Bardenet, Doucet, and Holmes]{bardenet2014towards}
R{\'e}mi Bardenet, Arnaud Doucet, and Chris Holmes.
\newblock Towards scaling up markov chain monte carlo: an adaptive subsampling approach.
\newblock In \emph{International conference on machine learning}, pages 405--413. PMLR, 2014.

\bibitem[Bardenet et~al.(2017)Bardenet, Doucet, and Holmes]{bardenet2017markov}
R{\'e}mi Bardenet, Arnaud Doucet, and Chris Holmes.
\newblock On markov chain monte carlo methods for tall data.
\newblock \emph{Journal of Machine Learning Research}, 18\penalty0 (47), 2017.

\bibitem[Baydin et~al.(2018)Baydin, Pearlmutter, Radul, and Siskind]{baydin2018automatic}
Atilim~Gunes Baydin, Barak~A Pearlmutter, Alexey~Andreyevich Radul, and Jeffrey~Mark Siskind.
\newblock Automatic differentiation in machine learning: a survey.
\newblock \emph{Journal of Marchine Learning Research}, 18:\penalty0 1--43, 2018.

\bibitem[Belouadah and Popescu(2020)]{scail_herding}
Eden Belouadah and Adrian Popescu.
\newblock Scail: Classifier weights scaling for class incremental learning.
\newblock In \emph{Proceedings of the IEEE/CVF winter conference on applications of computer vision}, pages 1266--1275, 2020.

\bibitem[Betancourt(2015)]{betancourt2015fundamental}
MJ~Betancourt.
\newblock The fundamental incompatibility of hamiltonian monte carlo and data subsampling.
\newblock \emph{arXiv preprint arXiv:1502.01510}, 2015.

\bibitem[Bierkens et~al.(2019)Bierkens, Fearnhead, and Roberts]{bierkens2019zig}
Joris Bierkens, Paul Fearnhead, and Gareth Roberts.
\newblock The zig-zag process and super-efficient sampling for bayesian analysis of big data.
\newblock \emph{The Annals of Statistics}, 47\penalty0 (3), jun 2019.
\newblock \doi{10.1214/18-aos1715}.
\newblock URL \url{https://doi.org/10.1214%2F18-aos1715}.

\bibitem[Bohdal et~al.(2020)Bohdal, Yang, and Hospedales]{LD}
Ondrej Bohdal, Yongxin Yang, and Timothy Hospedales.
\newblock Flexible dataset distillation: Learn labels instead of images.
\newblock \emph{arXiv preprint arXiv:2006.08572}, 2020.

\bibitem[Brier(1950)]{brier}
Glenn~W Brier.
\newblock Verification of forecasts expressed in terms of probability.
\newblock \emph{Monthly weather review}, 78\penalty0 (1):\penalty0 1--3, 1950.

\bibitem[Campbell and Beronov(2019)]{campbell_variational_bayesiancoreset}
Trevor Campbell and Boyan Beronov.
\newblock Sparse variational inference: Bayesian coresets from scratch.
\newblock \emph{Advances in Neural Information Processing Systems}, 32, 2019.

\bibitem[Campbell and Broderick(2018)]{campbell_tamara_greedy_2018}
Trevor Campbell and Tamara Broderick.
\newblock Bayesian coreset construction via greedy iterative geodesic ascent.
\newblock In \emph{International Conference on Machine Learning}, pages 698--706. PMLR, 2018.

\bibitem[Campbell and Broderick(2019)]{campbell_tamara_hilbert_2019}
Trevor Campbell and Tamara Broderick.
\newblock Automated scalable bayesian inference via hilbert coresets.
\newblock \emph{The Journal of Machine Learning Research}, 20\penalty0 (1):\penalty0 551--588, 2019.

\bibitem[Castro et~al.(2018)Castro, Mar{\'\i}n-Jim{\'e}nez, Guil, Schmid, and Alahari]{castro_herding}
Francisco~M Castro, Manuel~J Mar{\'\i}n-Jim{\'e}nez, Nicol{\'a}s Guil, Cordelia Schmid, and Karteek Alahari.
\newblock End-to-end incremental learning.
\newblock In \emph{Proceedings of the European conference on computer vision (ECCV)}, pages 233--248, 2018.

\bibitem[Cavalcante et~al.(2016)Cavalcante, Brasileiro, Souza, Nobrega, and Oliveira]{cavalcante2016computational}
Rodolfo~C Cavalcante, Rodrigo~C Brasileiro, Victor~LF Souza, Jarley~P Nobrega, and Adriano~LI Oliveira.
\newblock Computational intelligence and financial markets: A survey and future directions.
\newblock \emph{Expert Systems with Applications}, 55:\penalty0 194--211, 2016.

\bibitem[Cazenavette et~al.(2022)Cazenavette, Wang, Torralba, Efros, and Zhu]{MTT}
George Cazenavette, Tongzhou Wang, Antonio Torralba, Alexei~A Efros, and Jun-Yan Zhu.
\newblock Dataset distillation by matching training trajectories.
\newblock In \emph{Proceedings of the IEEE/CVF Conference on Computer Vision and Pattern Recognition}, pages 4750--4759, 2022.

\bibitem[Cazenavette et~al.(2023)Cazenavette, Wang, Torralba, Efros, and Zhu]{cazenavette2023generalizing}
George Cazenavette, Tongzhou Wang, Antonio Torralba, Alexei~A Efros, and Jun-Yan Zhu.
\newblock Generalizing dataset distillation via deep generative prior.
\newblock \emph{Proceedings of the IEEE/CVF conference on computer vision and pattern recognition}, 2023.

\bibitem[Chen et~al.(2022)Chen, Xu, and Campbell]{chen2022bayesian}
Naitong Chen, Zuheng Xu, and Trevor Campbell.
\newblock Bayesian inference via sparse hamiltonian flows.
\newblock \emph{Advances in Neural Information Processing Systems}, 35:\penalty0 20876--20888, 2022.

\bibitem[Chen et~al.(2014)Chen, Fox, and Guestrin]{sghmc}
Tianqi Chen, Emily Fox, and Carlos Guestrin.
\newblock Stochastic gradient hamiltonian monte carlo.
\newblock In \emph{International conference on machine learning}, pages 1683--1691. PMLR, 2014.

\bibitem[Chen et~al.(2012)Chen, Welling, and Smola]{Herding}
Yutian Chen, Max Welling, and Alex Smola.
\newblock Super-samples from kernel herding.
\newblock \emph{arXiv preprint arXiv:1203.3472}, 2012.

\bibitem[Croce et~al.(2021)Croce, Andriushchenko, Sehwag, Debenedetti, Flammarion, Chiang, Mittal, and Hein]{robustbench}
Francesco Croce, Maksym Andriushchenko, Vikash Sehwag, Edoardo Debenedetti, Nicolas Flammarion, Mung Chiang, Prateek Mittal, and Matthias Hein.
\newblock Robustbench: a standardized adversarial robustness benchmark.
\newblock In \emph{Thirty-fifth Conference on Neural Information Processing Systems Datasets and Benchmarks Track (Round 2)}, 2021.
\newblock URL \url{https://openreview.net/forum?id=SSKZPJCt7B}.

\bibitem[Cui et~al.(2022)Cui, Wang, Si, and Hsieh]{dc-bench}
Justin Cui, Ruochen Wang, Si~Si, and Cho-Jui Hsieh.
\newblock {DC}-{BENCH}: Dataset condensation benchmark.
\newblock In \emph{Thirty-sixth Conference on Neural Information Processing Systems Datasets and Benchmarks Track}, 2022.
\newblock URL \url{https://openreview.net/forum?id=Bs8iFQ7AM6}.

\bibitem[Deng and Russakovsky(2022)]{Rem_Past}
Zhiwei Deng and Olga Russakovsky.
\newblock Remember the past: Distilling datasets into addressable memories for neural networks.
\newblock \emph{arXiv preprint arXiv:2206.02916}, 2022.

\bibitem[Devlin et~al.(2018)Devlin, Chang, Lee, and Toutanova]{devlin2018bert}
Jacob Devlin, Ming-Wei Chang, Kenton Lee, and Kristina Toutanova.
\newblock Bert: Pre-training of deep bidirectional transformers for language understanding.
\newblock \emph{arXiv preprint arXiv:1810.04805}, 2018.

\bibitem[Dosovitskiy et~al.(2020)Dosovitskiy, Beyer, Kolesnikov, Weissenborn, Zhai, Unterthiner, Dehghani, Minderer, Heigold, Gelly, et~al.]{dosovitskiy2020image}
Alexey Dosovitskiy, Lucas Beyer, Alexander Kolesnikov, Dirk Weissenborn, Xiaohua Zhai, Thomas Unterthiner, Mostafa Dehghani, Matthias Minderer, Georg Heigold, Sylvain Gelly, et~al.
\newblock An image is worth 16x16 words: Transformers for image recognition at scale.
\newblock \emph{arXiv preprint arXiv:2010.11929}, 2020.

\bibitem[Du et~al.(2023)Du, Jiang, Tan, Zhou, and Li]{du2022minimizing}
Jiawei Du, Yidi Jiang, Vincent~TF Tan, Joey~Tianyi Zhou, and Haizhou Li.
\newblock Minimizing the accumulated trajectory error to improve dataset distillation.
\newblock \emph{Proceedings of the IEEE/CVF conference on computer vision and pattern recognition}, 2023.

\bibitem[Farahani and Hekmatfar(2009)]{farahani2009facility}
Reza~Zanjirani Farahani and Masoud Hekmatfar.
\newblock \emph{Facility location: concepts, models, algorithms and case studies}.
\newblock Springer Science \& Business Media, 2009.

\bibitem[Feldman and Langberg(2011)]{feldman2011unified}
Dan Feldman and Michael Langberg.
\newblock A unified framework for approximating and clustering data.
\newblock In \emph{Proceedings of the forty-third annual ACM symposium on Theory of computing}, pages 569--578, 2011.

\bibitem[Feldman et~al.(2011)Feldman, Faulkner, and Krause]{feldman2011scalable}
Dan Feldman, Matthew Faulkner, and Andreas Krause.
\newblock Scalable training of mixture models via coresets.
\newblock \emph{Advances in neural information processing systems}, 24, 2011.

\bibitem[Feldman et~al.(2020)Feldman, Schmidt, and Sohler]{feldman2020turning}
Dan Feldman, Melanie Schmidt, and Christian Sohler.
\newblock Turning big data into tiny data: Constant-size coresets for k-means, pca, and projective clustering.
\newblock \emph{SIAM Journal on Computing}, 49\penalty0 (3):\penalty0 601--657, 2020.

\bibitem[Frank et~al.(1956)Frank, Wolfe, et~al.]{frank1956algorithm}
Marguerite Frank, Philip Wolfe, et~al.
\newblock An algorithm for quadratic programming.
\newblock \emph{Naval research logistics quarterly}, 3\penalty0 (1-2):\penalty0 95--110, 1956.

\bibitem[Gawlikowski et~al.(2023)Gawlikowski, Tassi, Ali, Lee, Humt, Feng, Kruspe, Triebel, Jung, Roscher, et~al.]{gawlikowski2023survey}
Jakob Gawlikowski, Cedrique Rovile~Njieutcheu Tassi, Mohsin Ali, Jongseok Lee, Matthias Humt, Jianxiang Feng, Anna Kruspe, Rudolph Triebel, Peter Jung, Ribana Roscher, et~al.
\newblock A survey of uncertainty in deep neural networks.
\newblock \emph{Artificial Intelligence Review}, pages 1--77, 2023.

\bibitem[Gidaris and Komodakis(2018)]{convnet}
Spyros Gidaris and Nikos Komodakis.
\newblock Dynamic few-shot visual learning without forgetting.
\newblock In \emph{Proceedings of the IEEE conference on computer vision and pattern recognition}, pages 4367--4375, 2018.

\bibitem[Guo et~al.(2022)Guo, Zhao, and Bai]{guo2022deepcore}
Chengcheng Guo, Bo~Zhao, and Yanbing Bai.
\newblock Deepcore: A comprehensive library for coreset selection in deep learning.
\newblock In \emph{Database and Expert Systems Applications: 33rd International Conference, DEXA 2022, Vienna, Austria, August 22--24, 2022, Proceedings, Part I}, pages 181--195. Springer, 2022.

\bibitem[He et~al.(2016{\natexlab{a}})He, Zhang, Ren, and Sun]{he2016deep}
Kaiming He, Xiangyu Zhang, Shaoqing Ren, and Jian Sun.
\newblock Deep residual learning for image recognition.
\newblock In \emph{Proceedings of the IEEE conference on computer vision and pattern recognition}, pages 770--778, 2016{\natexlab{a}}.

\bibitem[He et~al.(2016{\natexlab{b}})He, Zhang, Ren, and Sun]{resnet}
Kaiming He, Xiangyu Zhang, Shaoqing Ren, and Jian Sun.
\newblock Deep residual learning for image recognition.
\newblock In \emph{Proceedings of the IEEE conference on computer vision and pattern recognition}, pages 770--778, 2016{\natexlab{b}}.

\bibitem[Hendrycks and Dietterich(2019)]{CIFAR10-C}
Dan Hendrycks and Thomas Dietterich.
\newblock Benchmarking neural network robustness to common corruptions and perturbations.
\newblock \emph{arXiv preprint arXiv:1903.12261}, 2019.

\bibitem[Hinton(2002)]{CD}
Geoffrey~E Hinton.
\newblock Training products of experts by minimizing contrastive divergence.
\newblock \emph{Neural computation}, 14\penalty0 (8):\penalty0 1771--1800, 2002.

\bibitem[Hoffman et~al.(2014)Hoffman, Gelman, et~al.]{hoffman2014no}
Matthew~D Hoffman, Andrew Gelman, et~al.
\newblock The no-u-turn sampler: adaptively setting path lengths in hamiltonian monte carlo.
\newblock \emph{J. Mach. Learn. Res.}, 15\penalty0 (1):\penalty0 1593--1623, 2014.

\bibitem[Huggins et~al.(2016)Huggins, Campbell, and Broderick]{huggins_tamara_logistic}
Jonathan Huggins, Trevor Campbell, and Tamara Broderick.
\newblock Coresets for scalable bayesian logistic regression.
\newblock \emph{Advances in neural information processing systems}, 29, 2016.

\bibitem[Jaakkola and Jordan(1997)]{jaakkola1997variational}
Tommi~S Jaakkola and Michael~I Jordan.
\newblock A variational approach to bayesian logistic regression models and their extensions.
\newblock In \emph{Sixth International Workshop on Artificial Intelligence and Statistics}, pages 283--294. PMLR, 1997.

\bibitem[Jiang et~al.(2022)Jiang, Gu, Liu, and Pan]{jiang2022delving}
Zixuan Jiang, Jiaqi Gu, Mingjie Liu, and David~Z Pan.
\newblock Delving into effective gradient matching for dataset condensation.
\newblock \emph{arXiv preprint arXiv:2208.00311}, 2022.

\bibitem[Johndrow et~al.(2020)Johndrow, Pillai, and Smith]{johndrow2020no}
James~E Johndrow, Natesh~S Pillai, and Aaron Smith.
\newblock No free lunch for approximate mcmc.
\newblock \emph{arXiv preprint arXiv:2010.12514}, 2020.

\bibitem[Jordan et~al.(1998)Jordan, Ghahramani, Jaakkola, and Saul]{jordan1998introduction}
Michael~I Jordan, Zoubin Ghahramani, Tommi~S Jaakkola, and Lawrence~K Saul.
\newblock An introduction to variational methods for graphical models.
\newblock \emph{Learning in graphical models}, pages 105--161, 1998.

\bibitem[Jordan et~al.(1999)Jordan, Ghahramani, Jaakkola, and Saul]{jordan1999introduction}
Michael~I Jordan, Zoubin Ghahramani, Tommi~S Jaakkola, and Lawrence~K Saul.
\newblock An introduction to variational methods for graphical models.
\newblock \emph{Machine learning}, 37:\penalty0 183--233, 1999.

\bibitem[Jospin et~al.(2022)Jospin, Laga, Boussaid, Buntine, and Bennamoun]{handsOnBayesian}
Laurent~Valentin Jospin, Hamid Laga, Farid Boussaid, Wray Buntine, and Mohammed Bennamoun.
\newblock Hands-on bayesian neural networks—a tutorial for deep learning users.
\newblock \emph{IEEE Computational Intelligence Magazine}, 17\penalty0 (2):\penalty0 29--48, 2022.

\bibitem[Kabir et~al.(2018)Kabir, Khosravi, Hosen, and Nahavandi]{kabir2018neural}
HM~Dipu Kabir, Abbas Khosravi, Mohammad~Anwar Hosen, and Saeid Nahavandi.
\newblock Neural network-based uncertainty quantification: A survey of methodologies and applications.
\newblock \emph{IEEE access}, 6:\penalty0 36218--36234, 2018.

\bibitem[Ker et~al.(2017)Ker, Wang, Rao, and Lim]{ker2017deep}
Justin Ker, Lipo Wang, Jai Rao, and Tchoyoson Lim.
\newblock Deep learning applications in medical image analysis.
\newblock \emph{Ieee Access}, 6:\penalty0 9375--9389, 2017.

\bibitem[Killamsetty et~al.(2021{\natexlab{a}})Killamsetty, Durga, Ramakrishnan, De, and Iyer]{killamsetty2021grad}
Krishnateja Killamsetty, Sivasubramanian Durga, Ganesh Ramakrishnan, Abir De, and Rishabh Iyer.
\newblock Grad-match: Gradient matching based data subset selection for efficient deep model training.
\newblock In \emph{International Conference on Machine Learning}, pages 5464--5474. PMLR, 2021{\natexlab{a}}.

\bibitem[Killamsetty et~al.(2021{\natexlab{b}})Killamsetty, Sivasubramanian, Ramakrishnan, and Iyer]{killamsetty2021glister}
Krishnateja Killamsetty, Durga Sivasubramanian, Ganesh Ramakrishnan, and Rishabh Iyer.
\newblock Glister: Generalization based data subset selection for efficient and robust learning.
\newblock In \emph{Proceedings of the AAAI Conference on Artificial Intelligence}, volume~35, pages 8110--8118, 2021{\natexlab{b}}.

\bibitem[Kim et~al.(2022{\natexlab{a}})Kim, Choi, Lee, Lee, Ha, and Lee]{BPC}
Balhae Kim, Jungwon Choi, Seanie Lee, Yoonho Lee, Jung-Woo Ha, and Juho Lee.
\newblock On divergence measures for bayesian pseudocoresets.
\newblock \emph{arXiv preprint arXiv:2210.06205}, 2022{\natexlab{a}}.

\bibitem[Kim et~al.(2022{\natexlab{b}})Kim, Kim, Oh, Yun, Song, Jeong, Ha, and Song]{kim2022dataset}
Jang-Hyun Kim, Jinuk Kim, Seong~Joon Oh, Sangdoo Yun, Hwanjun Song, Joonhyun Jeong, Jung-Woo Ha, and Hyun~Oh Song.
\newblock Dataset condensation via efficient synthetic-data parameterization.
\newblock In \emph{International Conference on Machine Learning}, pages 11102--11118. PMLR, 2022{\natexlab{b}}.

\bibitem[Korattikara et~al.(2014)Korattikara, Chen, and Welling]{korattikara2014austerity}
Anoop Korattikara, Yutian Chen, and Max Welling.
\newblock Austerity in mcmc land: Cutting the metropolis-hastings budget.
\newblock In \emph{International conference on machine learning}, pages 181--189. PMLR, 2014.

\bibitem[Krizhevsky and Hinton(2009)]{CIFAR10}
Alex Krizhevsky and Geoffrey Hinton.
\newblock Learning multiple layers of features from tiny images.
\newblock Technical Report~0, University of Toronto, Toronto, Ontario, 2009.
\newblock URL \url{https://www.cs.toronto.edu/~kriz/learning-features-2009-TR.pdf}.

\bibitem[Krizhevsky et~al.(2017{\natexlab{a}})Krizhevsky, Sutskever, and Hinton]{alexnet}
Alex Krizhevsky, Ilya Sutskever, and Geoffrey~E Hinton.
\newblock Imagenet classification with deep convolutional neural networks.
\newblock \emph{Communications of the ACM}, 60\penalty0 (6):\penalty0 84--90, 2017{\natexlab{a}}.

\bibitem[Krizhevsky et~al.(2017{\natexlab{b}})Krizhevsky, Sutskever, and Hinton]{krizhevsky2017imagenet}
Alex Krizhevsky, Ilya Sutskever, and Geoffrey~E Hinton.
\newblock Imagenet classification with deep convolutional neural networks.
\newblock \emph{Communications of the ACM}, 60\penalty0 (6):\penalty0 84--90, 2017{\natexlab{b}}.

\bibitem[Kucukelbir et~al.(2017)Kucukelbir, Tran, Ranganath, Gelman, and Blei]{kucukelbir2017automatic}
Alp Kucukelbir, Dustin Tran, Rajesh Ranganath, Andrew Gelman, and David~M Blei.
\newblock Automatic differentiation variational inference.
\newblock \emph{Journal of machine learning research}, 2017.

\bibitem[Le and Yang(2015)]{tinyimagenet}
Ya~Le and Xuan Yang.
\newblock Tiny imagenet visual recognition challenge.
\newblock \emph{CS 231N}, 7\penalty0 (7):\penalty0 3, 2015.

\bibitem[LeCun et~al.(1998)LeCun, Bottou, Bengio, and Haffner]{MNIST}
Yann LeCun, L{\'e}on Bottou, Yoshua Bengio, and Patrick Haffner.
\newblock Gradient-based learning applied to document recognition.
\newblock \emph{Proceedings of the IEEE}, 86\penalty0 (11):\penalty0 2278--2324, 1998.

\bibitem[Lee et~al.(2022)Lee, Chun, Jung, Yun, and Yoon]{DC_CS}
Saehyung Lee, Sanghyuk Chun, Sangwon Jung, Sangdoo Yun, and Sungroh Yoon.
\newblock Dataset condensation with contrastive signals.
\newblock In \emph{International Conference on Machine Learning}, pages 12352--12364. PMLR, 2022.

\bibitem[Li et~al.(2022)Li, Togo, Ogawa, and Haseyama]{DD_PP}
Guang Li, Ren Togo, Takahiro Ogawa, and Miki Haseyama.
\newblock Dataset distillation using parameter pruning.
\newblock \emph{arXiv preprint arXiv:2209.14609}, 2022.

\bibitem[Liu et~al.(2023)Liu, Gu, Wang, Zhu, Jiang, and You]{liu2023dream}
Yanqing Liu, Jianyang Gu, Kai Wang, Zheng Zhu, Wei Jiang, and Yang You.
\newblock Dream: Efficient dataset distillation by representative matching.
\newblock \emph{arXiv preprint arXiv:2302.14416}, 2023.

\bibitem[Loo et~al.(2022)Loo, Hasani, Amini, and Rus]{RFAD}
Noel Loo, Ramin Hasani, Alexander Amini, and Daniela Rus.
\newblock Efficient dataset distillation using random feature approximation.
\newblock \emph{arXiv preprint arXiv:2210.12067}, 2022.

\bibitem[Maclaurin and Adams(2014)]{maclaurin2014firefly}
Dougal Maclaurin and Ryan~P Adams.
\newblock Firefly monte carlo: Exact mcmc with subsets of data.
\newblock \emph{arXiv preprint arXiv:1403.5693}, 2014.

\bibitem[Manousakas et~al.(2020)Manousakas, Xu, Mascolo, and Campbell]{BPC_OG}
Dionysis Manousakas, Zuheng Xu, Cecilia Mascolo, and Trevor Campbell.
\newblock Bayesian pseudocoresets.
\newblock \emph{Advances in Neural Information Processing Systems}, 33:\penalty0 14950--14960, 2020.

\bibitem[Mirzasoleiman et~al.(2020)Mirzasoleiman, Bilmes, and Leskovec]{mirzasoleiman2020coresets}
Baharan Mirzasoleiman, Jeff Bilmes, and Jure Leskovec.
\newblock Coresets for data-efficient training of machine learning models.
\newblock In \emph{International Conference on Machine Learning}, pages 6950--6960. PMLR, 2020.

\bibitem[Naeini et~al.(2015)Naeini, Cooper, and Hauskrecht]{ECE}
Mahdi~Pakdaman Naeini, Gregory Cooper, and Milos Hauskrecht.
\newblock Obtaining well calibrated probabilities using bayesian binning.
\newblock In \emph{Proceedings of the AAAI conference on artificial intelligence}, volume~29, 2015.

\bibitem[Nagapetyan et~al.(2017)Nagapetyan, Duncan, Hasenclever, Vollmer, Szpruch, and Zygalakis]{nagapetyan2017true}
Tigran Nagapetyan, Andrew~B Duncan, Leonard Hasenclever, Sebastian~J Vollmer, Lukasz Szpruch, and Konstantinos Zygalakis.
\newblock The true cost of stochastic gradient langevin dynamics.
\newblock \emph{arXiv preprint arXiv:1706.02692}, 2017.

\bibitem[Naik et~al.(2022)Naik, Rousseau, and Campbell]{naik2022fast}
Cian Naik, Judith Rousseau, and Trevor Campbell.
\newblock Fast bayesian coresets via subsampling and quasi-newton refinement.
\newblock \emph{Advances in Neural Information Processing Systems}, 35:\penalty0 70--83, 2022.

\bibitem[Neal et~al.(2011{\natexlab{a}})Neal, Brooks, Gelman, Jones, Meng, et~al.]{neal2011handbook}
Radford~M Neal, S~Brooks, A~Gelman, G~Jones, XL~Meng, et~al.
\newblock Handbook of markov chain monte carlo.
\newblock \emph{Press C, editor}, 22011, 2011{\natexlab{a}}.

\bibitem[Neal et~al.(2011{\natexlab{b}})]{neal2011mcmc}
Radford~M Neal et~al.
\newblock Mcmc using hamiltonian dynamics.
\newblock \emph{Handbook of markov chain monte carlo}, 2\penalty0 (11):\penalty0 2, 2011{\natexlab{b}}.

\bibitem[Nguyen et~al.(2021)Nguyen, Chen, and Lee]{KIP}
Timothy Nguyen, Zhourong Chen, and Jaehoon Lee.
\newblock Dataset meta-learning from kernel ridge-regression.
\newblock In \emph{International Conference on Learning Representations}, 2021.
\newblock URL \url{https://openreview.net/forum?id=l-PrrQrK0QR}.

\bibitem[Pollock et~al.(2020)Pollock, Fearnhead, Johansen, and Roberts]{pollock2020quasi}
Murray Pollock, Paul Fearnhead, Adam~M Johansen, and Gareth~O Roberts.
\newblock Quasi-stationary monte carlo and the scale algorithm.
\newblock \emph{Journal of the Royal Statistical Society Series B: Statistical Methodology}, 82\penalty0 (5):\penalty0 1167--1221, 2020.

\bibitem[Radford et~al.(2021)Radford, Kim, Hallacy, Ramesh, Goh, Agarwal, Sastry, Askell, Mishkin, Clark, et~al.]{radford2021learning}
Alec Radford, Jong~Wook Kim, Chris Hallacy, Aditya Ramesh, Gabriel Goh, Sandhini Agarwal, Girish Sastry, Amanda Askell, Pamela Mishkin, Jack Clark, et~al.
\newblock Learning transferable visual models from natural language supervision.
\newblock In \emph{International conference on machine learning}, pages 8748--8763. PMLR, 2021.

\bibitem[Ranganath et~al.(2014)Ranganath, Gerrish, and Blei]{ranganath2014black}
Rajesh Ranganath, Sean Gerrish, and David Blei.
\newblock Black box variational inference.
\newblock In \emph{Artificial intelligence and statistics}, pages 814--822. PMLR, 2014.

\bibitem[Rebuffi et~al.(2017)Rebuffi, Kolesnikov, Sperl, and Lampert]{icarl_herding}
Sylvestre-Alvise Rebuffi, Alexander Kolesnikov, Georg Sperl, and Christoph~H Lampert.
\newblock icarl: Incremental classifier and representation learning.
\newblock In \emph{Proceedings of the IEEE conference on Computer Vision and Pattern Recognition}, pages 2001--2010, 2017.

\bibitem[Ritter et~al.(2018)Ritter, Botev, and Barber]{kfl}
Hippolyt Ritter, Aleksandar Botev, and David Barber.
\newblock A scalable laplace approximation for neural networks.
\newblock In \emph{6th international conference on learning representations, ICLR 2018-conference track proceedings}, volume~6. International Conference on Representation Learning, 2018.

\bibitem[Robert and Casella(2011)]{robert2011short}
Christian Robert and George Casella.
\newblock A short history of markov chain monte carlo: Subjective recollections from incomplete data.
\newblock \emph{Statistical Science}, 26\penalty0 (1), feb 2011.
\newblock \doi{10.1214/10-sts351}.
\newblock URL \url{https://doi.org/10.1214%2F10-sts351}.

\bibitem[Robert et~al.(1999)Robert, Casella, and Casella]{robert1999monte}
Christian~P Robert, George Casella, and George Casella.
\newblock \emph{Monte Carlo statistical methods}, volume~2.
\newblock Springer, 1999.

\bibitem[Sener and Savarese(2017)]{sener2017active}
Ozan Sener and Silvio Savarese.
\newblock Active learning for convolutional neural networks: A core-set approach.
\newblock \emph{arXiv preprint arXiv:1708.00489}, 2017.

\bibitem[Sermanet et~al.(2012)Sermanet, Chintala, and LeCun]{SVHN}
Pierre Sermanet, Soumith Chintala, and Yann LeCun.
\newblock Convolutional neural networks applied to house numbers digit classification.
\newblock In \emph{Proceedings of the 21st international conference on pattern recognition (ICPR2012)}, pages 3288--3291. IEEE, 2012.

\bibitem[Simonyan and Zisserman(2014)]{vgg}
Karen Simonyan and Andrew Zisserman.
\newblock Very deep convolutional networks for large-scale image recognition.
\newblock \emph{arXiv preprint arXiv:1409.1556}, 2014.

\bibitem[Szegedy et~al.(2013)Szegedy, Zaremba, Sutskever, Bruna, Erhan, Goodfellow, and Fergus]{szegedy2013intriguing}
Christian Szegedy, Wojciech Zaremba, Ilya Sutskever, Joan Bruna, Dumitru Erhan, Ian Goodfellow, and Rob Fergus.
\newblock Intriguing properties of neural networks.
\newblock \emph{arXiv preprint arXiv:1312.6199}, 2013.

\bibitem[Teh et~al.(2003)Teh, Welling, Osindero, and Hinton]{teh2003energy}
Yee~Whye Teh, Max Welling, Simon Osindero, and Geoffrey~E Hinton.
\newblock Energy-based models for sparse overcomplete representations.
\newblock \emph{Journal of Machine Learning Research}, 4\penalty0 (Dec):\penalty0 1235--1260, 2003.

\bibitem[Toneva et~al.(2018)Toneva, Sordoni, Combes, Trischler, Bengio, and Gordon]{Forgetting}
Mariya Toneva, Alessandro Sordoni, Remi Tachet~des Combes, Adam Trischler, Yoshua Bengio, and Geoffrey~J Gordon.
\newblock An empirical study of example forgetting during deep neural network learning.
\newblock \emph{arXiv preprint arXiv:1812.05159}, 2018.

\bibitem[Wainwright et~al.(2008)Wainwright, Jordan, et~al.]{wainwright2008graphical}
Martin~J Wainwright, Michael~I Jordan, et~al.
\newblock Graphical models, exponential families, and variational inference.
\newblock \emph{Foundations and Trends{\textregistered} in Machine Learning}, 1\penalty0 (1--2):\penalty0 1--305, 2008.

\bibitem[Wang et~al.(2022)Wang, Zhao, Peng, Zhu, Yang, Wang, Huang, Bilen, Wang, and You]{CAFE}
Kai Wang, Bo~Zhao, Xiangyu Peng, Zheng Zhu, Shuo Yang, Shuo Wang, Guan Huang, Hakan Bilen, Xinchao Wang, and Yang You.
\newblock Cafe: Learning to condense dataset by aligning features.
\newblock In \emph{Proceedings of the IEEE/CVF Conference on Computer Vision and Pattern Recognition}, pages 12196--12205, 2022.

\bibitem[Wang et~al.(2018)Wang, Zhu, Torralba, and Efros]{DD}
Tongzhou Wang, Jun-Yan Zhu, Antonio Torralba, and Alexei~A Efros.
\newblock Dataset distillation.
\newblock \emph{arXiv preprint arXiv:1811.10959}, 2018.

\bibitem[Welling(2009)]{welling2009herding}
Max Welling.
\newblock Herding dynamical weights to learn.
\newblock In \emph{Proceedings of the 26th Annual International Conference on Machine Learning}, pages 1121--1128, 2009.

\bibitem[Welling and Teh(2011)]{welling2011bayesian}
Max Welling and Yee~W Teh.
\newblock Bayesian learning via stochastic gradient langevin dynamics.
\newblock In \emph{Proceedings of the 28th international conference on machine learning (ICML-11)}, pages 681--688. Citeseer, 2011.

\bibitem[Xiao et~al.(2017)Xiao, Rasul, and Vollgraf]{FashionMNIST}
Han Xiao, Kashif Rasul, and Roland Vollgraf.
\newblock Fashion-mnist: a novel image dataset for benchmarking machine learning algorithms.
\newblock \emph{arXiv preprint arXiv:1708.07747}, 2017.

\bibitem[Yu et~al.(2023)Yu, Liu, and Wang]{DD_Survey}
Ruonan Yu, Songhua Liu, and Xinchao Wang.
\newblock Dataset distillation: A comprehensive review.
\newblock \emph{arXiv preprint arXiv:2301.07014}, 2023.

\bibitem[Zhang et~al.(2021)Zhang, Khanna, Kyrillidis, and Koyejo]{zhang_bayesiancoreset}
Jacky Zhang, Rajiv Khanna, Anastasios Kyrillidis, and Sanmi Koyejo.
\newblock Bayesian coresets: Revisiting the nonconvex optimization perspective.
\newblock In \emph{International Conference on Artificial Intelligence and Statistics}, pages 2782--2790. PMLR, 2021.

\bibitem[Zhang et~al.(2023)Zhang, Zhang, Lei, Mukherjee, Pan, Zhao, Ding, Li, and Xu]{zhang2022accelerating}
Lei Zhang, Jie Zhang, Bowen Lei, Subhabrata Mukherjee, Xiang Pan, Bo~Zhao, Caiwen Ding, Yao Li, and Dongkuan Xu.
\newblock Accelerating dataset distillation via model augmentation.
\newblock \emph{Proceedings of the IEEE/CVF conference on computer vision and pattern recognition}, 2023.

\bibitem[Zhao and Bilen(2021)]{DSA}
Bo~Zhao and Hakan Bilen.
\newblock Dataset condensation with differentiable siamese augmentation.
\newblock In \emph{International Conference on Machine Learning}, pages 12674--12685. PMLR, 2021.

\bibitem[Zhao and Bilen(2022)]{zhao2022synthesizing}
Bo~Zhao and Hakan Bilen.
\newblock Synthesizing informative training samples with gan.
\newblock \emph{arXiv preprint arXiv:2204.07513}, 2022.

\bibitem[Zhao and Bilen(2023)]{DM}
Bo~Zhao and Hakan Bilen.
\newblock Dataset condensation with distribution matching.
\newblock In \emph{Proceedings of the IEEE/CVF Winter Conference on Applications of Computer Vision}, pages 6514--6523, 2023.

\bibitem[Zhao et~al.(2021)Zhao, Mopuri, and Bilen]{GM}
Bo~Zhao, Konda~Reddy Mopuri, and Hakan Bilen.
\newblock Dataset condensation with gradient matching.
\newblock In \emph{International Conference on Learning Representations}, 2021.
\newblock URL \url{https://openreview.net/forum?id=mSAKhLYLSsl}.

\bibitem[Zhao et~al.(2023)Zhao, Li, Qin, and Yu]{Improve_DM}
Ganlong Zhao, Guanbin Li, Yipeng Qin, and Yizhou Yu.
\newblock Improved distribution matching for dataset condensation.
\newblock In \emph{Proceedings of the IEEE/CVF conference on computer vision and pattern recognition}, 2023.

\bibitem[Zhou et~al.(2022)Zhou, Nezhadarya, and Ba]{FREPO}
Yongchao Zhou, Ehsan Nezhadarya, and Jimmy Ba.
\newblock Dataset distillation using neural feature regression.
\newblock In Alice~H. Oh, Alekh Agarwal, Danielle Belgrave, and Kyunghyun Cho, editors, \emph{Advances in Neural Information Processing Systems}, 2022.
\newblock URL \url{https://openreview.net/forum?id=2clwrA2tfik}.

\end{thebibliography}

\newpage

\onecolumn

\title{Bayesian Pseudo-Coresets via Contrastive Divergence\\(Supplementary Material)}
\maketitle
\appendix

\section{Training details and Hyper parameters}
\label{sec:app-training-details}
In this section, we provide the implementation details of the proposed method. Our implementation can be found \href{https://github.com/backpropagator/BPC-CD}{https://github.com/backpropagator/BPC-CD}. During the training process, we randomly initialize the synthetic dataset using samples from the original training set. The overall cardinality of these synthetic sets is determined by the number of images considered for every class (ipc). For our experiments, we have considered ipc values of 1, 10 and 50. Furthermore, similar to previous works~\cite{BPC,MTT}, we have used Differentiable Siamese Augmentation (DSA)~\citep{DSA} strategies to enhance the performance of our model. DSA strategies include random crop, random flip, random brightness, random scale, and rotation. At any instant, we apply one of these augmentations to train our network. These augmentation techniques ensure that the model does not overfit on the given synthetic set and generates optimal parameters. DSA is applied to the synthetic set while running langevin dynamics and calculating the contrastive-divergence-based loss function. 

We have also conducted additional experiments to evaluate the effectiveness of our method and other BPC baselines in the absence of DSA. The findings of the experiment are reported in Table~\ref{tab:dsa_comparison}. The results clearly indicate that the DSA  has a positive impact on the performance of all BPC methods, which align with the observation made by~\citet{dc-bench}. Nevertheless, even without the DSA-based augmentation strategy, our method outperforms other BPC baselines for 1 ipc.

As for the network used to calculate the energy, we take inspiration from previous works~\citep{BPC,DD,DC_CS,MTT} and use a ConvNet architecture~\citep{convnet}. This architecture consists of multiple blocks of convolutional layer with filter dimension of  $3\times3$ and channel size of $128$. The network uses instance normalization, maxpool layer with stride $2$, and RELU activation. In our experiments, we have used an architecture with three such blocks of convolution layers.

Next, we create a buffer of trajectories to sample parameters from the posterior of the original dataset. For this, we  generate $100$ different trajectories, each with $50$ epochs trained using SGD optimizer with a batch size of $256$ on the original training set. These parameters are used to obtain the gaussian variational approximation to estimate the loss function. Further, we use diagonal covariance matrix with diagonal entry of $0.001$ for the re-parameterization trick used in gaussian approximation.

The pseudocode of the implementation is presented in Algorithm~\ref{alg:the_alg}. The hyperparameters used are as follows: $P = 2000$, $\lambda = 0.01$, $n = 50$, $L = 100$, $\Sigma_{\rvx}^{1/2} = 0.001I$. These hyper-parameters are fixed across all the datasets. Further, we observe that there $\gamma$ that can be fine-tuned for marginal improvements in performance. Specifically, $\gamma$ is varied between $\{1, 10, 100, 1000\}$. Note that, we use same set of parameter to sample parameters during inference.
All the experiments are conducted on a single NVIDIA RTX A6000 GPUs with 48GB memory.

\begin{table}[!h]
\centering
\caption{Comparison of the proposed method against other BPC baselines without using DSA on CIFAR10 dataset.}
\label{tab:dsa_comparison}
\resizebox{\columnwidth}{!}{%
\begin{tabular}{rr|rr|rr|rr|rr}
\hline
\multicolumn{2}{c|}{\textbf{BPC-rkl (sghmc)}} & \multicolumn{2}{c|}{\textbf{BPC-W (sghmc)}} & \multicolumn{2}{c|}{\textbf{BPC-fkl (hmc)}} & \multicolumn{2}{c|}{\textbf{BPC-fkl (sghmc)}} & \multicolumn{2}{c}{\textbf{Ours}}                    \\ \hline
\multicolumn{1}{c}{ipc =1}                 & \multicolumn{1}{c|}{ipc = 10}              & \multicolumn{1}{c}{ipc =1}               & \multicolumn{1}{c|}{ipc = 10}             & \multicolumn{1}{c}{ipc =1}               & \multicolumn{1}{c|}{ipc = 10}             & \multicolumn{1}{c}{ipc =1}                & \multicolumn{1}{c|}{ipc = 10}              & \multicolumn{1}{c}{ipc =1}                    & \multicolumn{1}{c}{ipc = 10}                  \\ \hline
$19.70 \pm 1.06$       & $36.41 \pm 0.75$      & $27.66 \pm 0.8$      & $39.61 \pm 1.12$     & $32.61 \pm 1.50$     & $38.12 \pm 1.19$     & $28.25 \pm 0.92$      & \boldsymbol{$41.85 \pm 1.47$}      & \boldsymbol{$34.94 \pm 0.72$} & $41.02 \pm 0.66$ \\ \hline
\end{tabular}%
}
\end{table}

\begin{algorithm}
\caption{Proposed Algorithm }
\label{alg:the_alg}
\textbf{Input :} Set of SGD trajectories obtained from original dataset $(\tau)$, Number of langevin steps ($L$) needed to sample parameter from $\pi_{\td{\rvx}}$, Langevin step size  ($\lambda$), Step size to modify pseudo-coreset ($\gamma$), Number of epochs (P)  \\
\begin{algorithmic}[1]
\State Initialize pseudo-coreset ($\td{x}$) using samples from original dataset $x$.

\For {step in [1... P] : }
\State Sample  $\tau_i \sim \tau$ 
\State Sample $\theta_k^+ \sim \tau_i$ where $\theta_k^+$ are parameters associated with $k^{th}$ epoch for $i^{th}$ trajectory.
\State Let $\theta^+$ = $\theta_{k}^{+} + \Sigma_{\rvx}^{1/2}\varepsilon_{\rvx}$, $\varepsilon \sim \gN(0,I)$
\State Let $\theta^{-}_{0}  = \theta^{+} $
\For {t in [0 .... L] :  }
    \State Calculate energy associated with $\td{\rvx}$ and parameter $\theta^{-}_{t}$ i.e. $E(\theta^{-}_{t},\td{\rvx})$)
    
    \State  $\theta^{-}_{t+1} = \theta^{-}_{t} - \lambda(\nabla_{\theta}E(\theta^{-}_{t},\td{\rvx})) + \eta \text{,}\hspace{1.5mm} \eta \sim {\gN}(0,I)$

\EndFor
 \State Let $\theta^-$ = $\theta^-_{L}$
 \State Calculate  $ \gL = E((\theta^{+},\td{\rvx})) - E({(\theta^{-} , \td{\rvx} )}) $
 \State $ \td{\rvx} \gets \td{\rvx} - \gamma \nabla_{\td{\rvx}}  \gL $
 \EndFor

\end{algorithmic}
\end{algorithm}

\section{Experimental Setup}

{\subsection{Baseline Setup}
We primarily present the results for different BPC frameworks. The experiment of Table~\ref{sota-comp-BPC} in the main manuscript uses the original hyperparameters mentioned in the respective papers. In cases where hyperparameters were not explicitly specified, we employed the default hyperparameters of CIFAR10. 
We have presented the results for BPC methods with only 1 and 10 ipc for the CIFAR100 and T-ImageNet datasets. We could not report the result for other scenarios due to the computational limitations. These methods demand a significant amount of GPU memory, which we currently lack, making it impractical to compute the desired results.

\subsection{GPU and Time Consumption}
We assess the computational efficiency of our method relative to other baselines by comparing the GPU memory consumption and the training time required to generate the pseudo-coresets for a single iteration. The findings of our results are presented in Fig.~\ref{gpu-time}, where the iteration time is calculated by taking the average of the total time for 100 different iterations.

As illustrated in Fig.~\ref{gpu-time}, our method requires relatively less time compared to other BPC methods for low ipc values and outperforms BPC-W for higher ipc values. Additionally, in our examination of GPU memory usage, we observe that BPC-W shows linear scaling in GPU memory consumption as the number of images per class increases. In contrast, our method maintains consistent memory usage across all ipc values. In our experiment, we found that our method utilizes only 37GB of memory, even for higher images per class. It's worth noting that other BPC baselines such as BPC-fKL and BPC-rKL are more memory-efficient than our method and deliver consistent performance across all the ipc values. We attribute this observation to the fact that BPC-fkl and BPC-rkl avoid MCMC sampling during training by making use of relevant approximations. Whereas, our method makes use of gradient-based MCMC sampling (langevin dynamics) for estimation of the objective function. For this reason, the GPU consumption of the proposed method is relatively higher than that of BPC-fkl and BPC-rkl.

Further, for a pseudo-coreset of size $m$, we run langevin-dynamics for $k$ steps which leads to a complexity of $\mathcal{O}(km)$. Compared to inference on entire dataset of size $n$, the same would have a complexity of $\mathcal{O}(kn)$. Since $m\ll n$, pseudo-coresets are much more efficient for inference compared to full data. Since the pseudo-coreset size ($m$) is fixed during inference and only final value of each iteration is required in the consequent iteration of langevin dynamics, the memory complexity is just $\mathcal{O}(m)$. Again since $m\ll n$, pseudo-coresets are efficient in term of memory as well compared to full data inference.


\begin{figure*}[!t]
\centering
   \begin{subfigure}[t]{0.49\textwidth}
   \includegraphics[keepaspectratio, width=\textwidth]{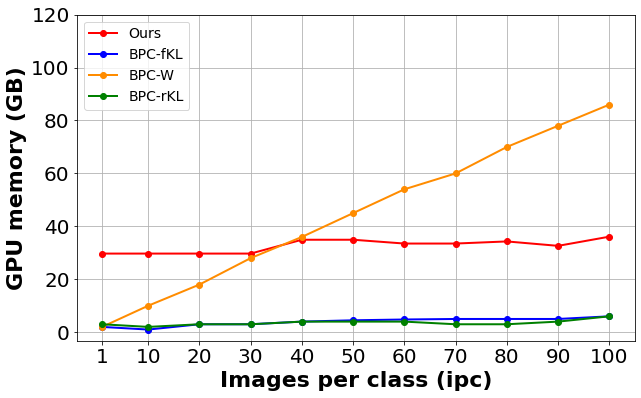}
   \caption{\label{fig:gpu-mem}}
   \end{subfigure}
~
    \begin{subfigure}[t]{0.49\textwidth}
   \includegraphics[keepaspectratio, width=\textwidth]{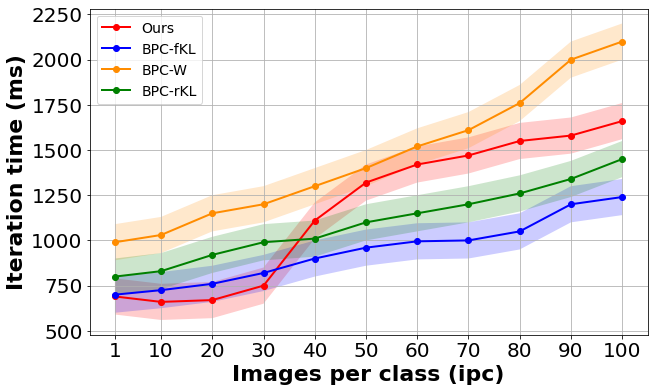}
   \caption{\label{fig:gpu-iter}}
    \end{subfigure}
\caption{\label{gpu-time}Computing GPU memory costs along with training time for different image per class.}
\end{figure*}

\section{Comparison with coreset  methods}
We have conducted a comparative analysis of our method with other coreset techniques such as Herding~\citep{Herding}, K-Center~\citep{sener2017active}, and Forgetting~\citep{Forgetting}. The outcomes of our experiments are listed in Table~\ref{tab:sota-comp-coreset}. The result clearly shows that our method outperforms other coreset techniques on all the dataset.   
\begin{table*}[ht!]
\centering
  \caption{\label{tab:sota-comp-coreset}Comparison of the proposed method with coreset baselines. The results are noted in form of (mean $\pm$ std. dev) where we have obtained test accuracy over five independent runs on the pseudo-coreset. The best performer across all methods is denoted in bold ($\boldsymbol{x\pm s}$). For ease of comparison, we color the second best performer with \textcolor{blue}{blue} color.}
\resizebox{0.75\textwidth}{!}
 {
 \renewcommand{\arraystretch}{1.1}
\begin{tabular}{l|rr|rrr|r}
 \hline
 &
   \multicolumn{1}{c}{\textbf{ipc}} &
   \multicolumn{1}{c|}{\textbf{Ratio(\%)}} &
   \multicolumn{1}{c}{\textbf{Herding}} &
   \multicolumn{1}{c}{\textbf{K-Center}} &
   \multicolumn{1}{c|}{\textbf{Forgetting}} &
   \multicolumn{1}{c}{\textbf{Ours}} 
   \\ \hline
 &
  1 &
  0.017 &
  $89.2 \pm 1.6$ &
  {\color{blue}{$89.3 \pm 1.5$ }}&
  $35.5 \pm 5.6$ &
 \boldsymbol{$93.42 \pm 0.09$} 
   \\
 &
  10 &
  0.17 &
  {\color{blue}{${93.7} \pm 0.3$}} &
  $84.4 \pm 1.7$ &
  $68.1 \pm 3.3$ &
  \boldsymbol{$97.71 \pm 0.24$} 
   \\
 \textbf{\multirow{-3}{*}{\textbf{MNIST}}} &
  50 &
  0.83 &
  $94.8 \pm 0.2$ &
  {\color{blue}{${97.4} \pm 0.3$}} &
  $88.2 \pm 1.2$ &
  \boldsymbol{$98.91 \pm 0.22$} 
   \\ \hline
 &
  1 &
  0.017 &
  {\color{blue}{${67.0} \pm 1.9$}} &
  $66.9 \pm 1.8$ &
  $42.0 \pm 5.5$ &
   \boldsymbol{$77.29 \pm 0.50$}
   \\
 &
  10 &
  0.17 &
  {\color{blue}{${71.1} \pm 0.7$}} &
  $54.7 \pm 1.5$ &
  $53.9 \pm 2.0$ &
  \boldsymbol{$88.40 \pm 0.21$} 
   \\
\multirow{-3}{*}{\textbf{FMNIST}} &
  50 &
  0.83 &
  {\color{blue}{${71.9} \pm 0.8$}} &
  $68.3 \pm 0.8$ &
  $55.0 \pm 1.1$ &
  \boldsymbol{$89.47 \pm 0.06$} 
   \\ \hline
 &
  1 &
  0.014 &
  $20.9 \pm 1.3$ &
  {\color{blue}{${21.0} \pm 1.5$}} &
  $12.1 \pm 1.7$ &
  \boldsymbol{$66.74 \pm 0.09$} 
   \\
 &
  10 &
  0.14 &
  {\color{blue}{${50.5} \pm 3.3$}} &
  $14.0 \pm 1.3$ &
  $16.8 \pm 1.2$ &
  \boldsymbol{$82.32 \pm 0.56$} 
   \\
 \multirow{-3}{*}{\textbf{SVHN}}&
  50 &
  0.7 &
  {\color{blue}{${72.6} \pm 0.8$}} &
  $20.1 \pm 1.4$ &
  $27.2 \pm 1.5$ &
  \boldsymbol{$88.41 \pm 0.12$} 
   \\ \hline
 &
  1 &
  0.02 &
  $21.5 \pm 1.2$ &
  {\color{blue}{${21.5} \pm 1.3$}} &
  $13.5 \pm 1.2$ &
  \boldsymbol{$46.87 \pm 0.20$} 
   \\
 &
  10 &
  0.2 &
  {\color{blue}{${31.6} \pm 0.7$}} &
  $14.7 \pm 0.9$ &
  $23.3 \pm 1.0$ &
  \boldsymbol{$56.39 \pm 0.70$} 
   \\
\multirow{-3}{*}{\textbf{Cifar10}} &
  50 &
  1 &
  $23.3 \pm 1.0$ &
  {\color{blue}{${27.0} \pm 1.4$}} &
  $23.3 \pm 1.1$ &
  \boldsymbol{$71.93 \pm 0.17$} 
   \\ \hline
 &
  1 &
  0.2 &
  {\color{blue}{${8.4} \pm 0.3$}} &
  $8.3 \pm 0.3$ &
  $4.5 \pm  0.2$ &
  \boldsymbol{$23.97 \pm 0.11$} 
   \\
\multirow{-2}{*}{\textbf{Cifar100}} &
  10 &
  2 &
  {\color{blue}{${17.3} \pm 0.3$}} &
  $7.1 \pm 0.2$ &
  $15.1 \pm 0.3$ &
  \boldsymbol{$28.42 \pm 0.24$} 
   \\ \hline
 &
  1 &
  0.2 &
  $2.8 \pm 0.2$ &
  {\color{blue}{${3.03} \pm 0.1$}} &
  $1.6 \pm 0.1$ &
  \boldsymbol{$8.39 \pm 0.07$} 
   \\
\multirow{-2}{*}{\textbf{T-ImageNet}} &
  10 &
  2 &
  $6.3 \pm 0.2$ &
  {\color{blue}{${11.38} \pm 0.1$}} &
  $5.1 \pm 0.2$ &
  \boldsymbol{$17.82 \pm 0.39$} 
   \\ \hline
  
\end{tabular}
}
\end{table*}

\section{Comparison with dataset condensation techniques}

\begin{table*}[!t]
    \centering
    \caption{Efficiency comparison of the proposed method with several dataset condensation methods on CIFAR10.}
    \resizebox{\textwidth}{!}{
            \begin{tabular}{l|rr|rr|rr|rr}
\toprule
\multirow{2}{*}{\textbf{ipc}} & \multicolumn{2}{c|}{\textbf{GM}}                                                      & \multicolumn{2}{c|}{\textbf{DSA}}                                                     & \multicolumn{2}{c|}{\textbf{MTT}}                                                     & \multicolumn{2}{c}{\textbf{Ours}}                                                    \\
                              & \multicolumn{1}{c}{\textbf{GPU Memory (GB)}} & \multicolumn{1}{c|}{\textbf{Time (s)}} & \multicolumn{1}{c}{\textbf{GPU Memory (GB)}} & \multicolumn{1}{c|}{\textbf{Time (s)}} & \multicolumn{1}{c}{\textbf{GPU Memory (GB)}} & \multicolumn{1}{c|}{\textbf{Time (s)}} & \multicolumn{1}{c}{\textbf{GPU Memory (GB)}} & \multicolumn{1}{c}{\textbf{Time (s)}} \\ \midrule
\textbf{1}                    & $38.88$                                      & $0.4$                                  & $43.09$                                      & $0.3$                                 & $44.82$                                      & $1.2$                                  & $\mathbf{37.10}$                                      & $\mathbf{0.71}$                                 \\
\textbf{10}                   & $42.78$                                      & $11.4$                                 & $46.06$                                      & $7.5$                                  & $50.98$                                      & $12.7$                                 & $\mathbf{38.28}$                                      & $\mathbf{0.75}$                                 \\
\textbf{50}                   & $46.39$                                      & $32.9$                                 & $53.20$                                       & $24.4$                                 & $68.10$                                      & $26.8$                                 & $\mathbf{38.44}$                                      & $\mathbf{1.26}$                                 \\ \bottomrule
\end{tabular}%
}
\label{tab:eff-comp-dc}
\end{table*}

 \begin{table*}[!ht]

  \caption{\label{sota-comp-condense}Comparison of the proposed method with dataset-condensation baselines. The results are noted in form of (mean $\pm$ std. dev) where we have obtained test accuracy over five independent runs on the pseudo-coreset. The best performer across all methods is denoted in bold ($\boldsymbol{x\pm s}$). For ease of comparison, we color the second best performer with \textcolor{blue}{blue} color.}
 \resizebox{\textwidth}{!}
 {
 \renewcommand{\arraystretch}{1.1}
\begin{tabular}{l|rr|rrrrrrrrr|r}
\hline
 &
  \multicolumn{1}{c}{\textbf{Img/cls}} &
  \multicolumn{1}{c|}{\textbf{Ratio\%}} &
  \multicolumn{1}{c}{\textbf{DD}} &
  \multicolumn{1}{c}{\textbf{LD}} &
  \multicolumn{1}{c}{\textbf{GM}} &
  \multicolumn{1}{c}{\textbf{DSA}} &
  \multicolumn{1}{c}{\textbf{DM}} &
  \multicolumn{1}{c}{\textbf{CAFE}} &
  \multicolumn{1}{c}{\textbf{CAFE+DSA}} &
  \multicolumn{1}{c}{\textbf{KIP}} &
  \multicolumn{1}{c|}{\textbf{MTT}} &
  \multicolumn{1}{c}{\textbf{Ours}} \\ \hline
 &
  1 &
  0.017 &
  - &
  $60.60 \pm 2.86$ &
  $92.01 \pm 0.25$ &
  $87.60 \pm 0.07$ &
  $88.89 \pm 0.57$ &
  {\color{blue}{${93.10} \pm 0.30$}} &
  $90.80 \pm 0.50$ &
  $85.46 \pm 0.04$ &
  $89.85 \pm 0.01$ &
  \boldsymbol{$93.42 \pm 0.09$} \\
 &
  10 &
  0.17 &
  $79.71 \pm 8.3$ &
  $87.05 \pm 0.50$ &
  $97.58 \pm 0.10$ &
  $97.39 \pm 0.06$ &
  $96.58 \pm 0.11$ &
  $97.20 \pm 0.20$ &
  $97.50 \pm 0.10$ &
  $97.15 \pm 0.11$ &
  {\color{blue}{${97.70} \pm 0.02$}} &
  \boldsymbol{$97.71 \pm 0.24$} \\
\multirow{-3}{*}{\textbf{MNIST}} &
  50 &
  0.83 &
  - &
  $93.30 \pm 0.30$ &
  $98.81 \pm 0.03$ &
  $98.97 \pm 0.04$ &
  $98.22 \pm 0.05$ &
  $98.60 \pm 0.20$ &
  {\color{blue}{${98.90} \pm 0.20$}} &
  $98.36 \pm 0.08$ &
  $98.6 \pm 0.01$ &
  \boldsymbol{$98.91 \pm 0.22$} \\ \hline
 &
  1 &
  0.017 &
  - &
  - &
  $70.83 \pm 0.01$ &
  $70.45 \pm 0.57$ &
  $71.92 \pm 0.70$ &
  $77.10 \pm 0.90$ &
  $73.70 \pm 0.70$ &
  - &
  {\color{blue}{${77.14} \pm 0.01$}} &
  \boldsymbol{$77.29 \pm 0.50$} \\
 &
  10 &
  0.17 &
  - &
  - &
  $81.93 \pm 0.07$ &
  $84.70 \pm 0.11$ &
  $83.25 \pm 0.09$ &
  $83.00 \pm 0.40$ &
  $83.00 \pm 0.30$ &
  - &
  \boldsymbol{${88.76} \pm 0.01$} &
  {\color{blue}{$88.40 \pm 0.21$}} \\
\multirow{-3}{*}{\textbf{FMNIST}} &
  50 &
  0.83 &
  - &
  - &
  $83.26 \pm 0.17$ &
  $88.55 \pm 0.56$ &
  $87.65 \pm 0.03$ &
  $84.80 \pm 0.40$ &
  $88.20 \pm 0.30$ &
  - &
  {\color{blue}{${89.33} \pm 0.15$}} &
  \boldsymbol{$89.47 \pm 0.06$} \\ \hline
 &
  1 &
  0.014 &
  - &
  - &
  $30.49 \pm 0.57$ &
  $31.18 \pm 0.43$ &
  $19.25 \pm 1.39$ &
  $42.60 \pm 3.30$ &
  $42.90 \pm 3.01$ &
  - &
  {\color{blue}{${57.55} \pm 0.02$}} &
  \boldsymbol{$66.74 \pm 0.09$} \\
 &
  10 &
  0.14 &
  - &
  - &
  $75.10 \pm 0.40$ &
  $78.39 \pm 0.3$ &
  $71.42 \pm 1.01$ &
  $75.90 \pm 0.60$ &
  {\color{blue}{${77.90} \pm 0.60$}} &
  - &
  $72.56 \pm 0.01$ &
  \boldsymbol{$82.32 \pm 0.56$} \\
\multirow{-3}{*}{\textbf{SVHN}} &
  50 &
  0.7 &
  - &
  - &
  $81.70 \pm 0.14$ &
  $82.50 \pm 0.34$ &
  $82.41 \pm0.52$ &
  $81.30 \pm 0.30$ &
  $82.30 \pm 0.40$ &
  - &
  {\color{blue}{${83.73} \pm 0.33$}} &
  \boldsymbol{$88.41 \pm 0.12$} \\ \hline
 &
  1 &
  0.02 &
  - &
  $25.38 \pm 0.2$ &
  $28.10 \pm0.56$ &
  $29.01 \pm 0.64$ &
  $26.40 \pm 0.42$ &
  $30.30 \pm 1.10$ &
  $31.60 \pm 0.80$ &
  $40.50 \pm 0.40$ &
  {\color{blue}{${46.08} \pm 0.80$}} &
  \boldsymbol{$46.87 \pm 0.2$} \\
 &
  10 &
  0.2 &
  $39.14 \pm 2.30$ &
  $37.50 \pm 0.60$ &
  $44.14 \pm 0.60$ &
  $51.85 \pm 0.43$ &
  $48.66 \pm 0.03$ &
  $46.30 \pm 0.60$ &
  $50.90 \pm 0.50$ &
  $53.10 \pm 0.50$ &
  \boldsymbol{${64.27} \pm 0.80$} &
  {\color{blue}{$56.39 \pm 0.70$}} \\
\multirow{-3}{*}{\textbf{CIFAR10}} &
  50 &
  1 &
  - &
  $41.70 \pm 0.50$ &
  $53.73 \pm 0.44$ &
  $60.77 \pm 0.45$ &
  $62.70 \pm 0.07$ &
  $55.50 \pm 0.60$ &
  $63.30 \pm 0.40$ &
  $58.60 \pm 0.40$ &
  {\color{blue}{${71.26} \pm 0.50$}} &
  \boldsymbol{$71.93 \pm 0.17$} \\ \hline
 &
  1 &
  0.2 &
  - &
  $11.50 \pm 0.40$ &
  $12.65 \pm 0.32$ &
  $13.88 \pm 0.29$ &
  $11.35 \pm 0.18$ &
  $12.04 \pm 0.01$ &
  $12.90 \pm 0.30$ &
  $14.01 \pm 0.30$ &
  {\color{blue}{${23.62} \pm 0.63$}} &
  \boldsymbol{$23.97 \pm 0.11$} \\
\multirow{-2}{*}{\textbf{CIFAR100}} &
  10 &
  2 &
  - &
  - &
  $25.28 \pm 0.29$ &
  $32.34 \pm 0.40$ &
  $29.38 \pm 0.26$ &
  $29.04 \pm 0.01$ &
  $27.80 \pm 0.30$ &
  {\color{blue}{$31.50 \pm 0.20$}} &
  \boldsymbol{${36.96} \pm 0.15$} &
  $28.42 \pm 0.24$ \\ \hline
 &
  1 &
  0.2 &
  - &
  - &
  $5.27 \pm 0.01$ &
  $5.67 \pm 0.01$ &
  $3.82 \pm 0.01$ &
  - &
  - &
  - &
  {\color{blue}{${8.27} \pm 0.01$}} &
  \boldsymbol{$8.39 \pm 0.07$} \\
\multirow{-2}{*}{\textbf{T-ImageNet}} &
  10 &
  2 &
  - &
  - &
  $12.83 \pm 0.01$ &
  $16.43 \pm 0.02$ &
  $13.51 \pm 0.01$ &
  - &
  - &
  - &
  \boldsymbol{${20.11} \pm 0.02$} &
  {\color{blue}{$17.82 \pm 0.39$}} \\ \hline
\end{tabular}
 }
 \end{table*}

We also compare our method with other data condensation (DC) techniques like Distillation (DD)~\citep{DD}, Flexible Dataset Distillation (LD)~\citep{LD}, Gradient Matching (DC)~\citep{GM}, Differentiable Siamese Augmentation (DSA)~\citep{DSA}, Distribution Matching (DM)~\citep{DM}, Neural Ridge Regression (KIP)~\citep{KIP}, Condensed data to align features (CAFE)~\citep{CAFE} and Matching Training Trajectories (MTT)~\citep{MTT}. 
\par There are few fundamental differences between dataset condensation methods and BPC methods. We first enumerate these differences here for clarity:
\begin{enumerate}
    \item \textbf{Objective function}: One of the key differences between Dataset Condensation (DC) and BPC is the underlying formulation and the loss objective. Dataset condensation methods mostly rely on heurist objective function to `\textbf{match the performance}' of synthetic data and original data; where the measure of performance matching varies for different methods. For e.g., GM~\citep{GM} relies on matching the gradient direction of a model trained on synthetic data with a model trained on the original data; similarly, MTT~\citep{MTT} relies on matching SGD trajectories of the synthetic data and original data. Hence, these methods don't have a principled way of coming up with loss objective. However, BPC methods on the other hand follow a single principle for loss objective - a divergence measure between true posterior and pseudo-coreset posterior, which is much more principled. Different BPC method opt for different divergence measure, in context of our work, we choose to work with contrastive divergence for the reasons outlined in the paper.
    \item \textbf{Bayesian v/s Non-Bayesian}: Another fundamental difference between the two domains is that BPC methods are purely bayesian, particularly, they treat parameters of the network as random variable and work on matching the distribution of this random variable via divergence minimization. On the other hand, DC methods are non-bayesian and work with point estimates of models.
    \item \textbf{Optimization Strategy}: Finally, the pivotal difference between BPC methods and DC methods, that makes the former more efficient is the optimization strategy employed. Since DC methods rely on `performance matching', it is inadvertent to train a model on synthetic and original data, then match the performance using appropriate metric. To learn the synthetic data in this case, one has to backpropagate the gradients through the model training steps as well. This would require computation of second-order derivatives. This is what we refer to as bi-level optimization. However, in case of BPC methods like ours, there is no such need of higher order derivatives. This is because we don't use bi-level optimization for BPC construction. This difference in optimization strategy makes BPC much more efficient in practice. The quantitative results to illustrate this claim can be found in Table~\ref{tab:eff-comp-dc}, where we compare the GPU memory consumed and time taken for every iteration by different methods. It can be seen that dataset condensation consume significant memory and time. 
\end{enumerate}
\par The performance comparison results are shown in Table~\ref{sota-comp-condense}. We find that the performance of our method is better than almost all the DC baselines, whereas MTT stands out to be a close second in most of the cases. This shows that our method, although falling under the category of Bayesian pseudo-coreset, achieves a performance that is comparable to that of heuristic DC methods. It is to be noted that the DC methods are not the direct competitors of our method. However, we have shown that our method, although a BPC, surpasses (or comes very close to) the SoTA DC methods such as MTT~\citep{MTT}.

\section{Visualizations for CIFAR100 and Tiny-ImageNet}
In this section, we present the visualizations for pseudo-coresets of large datasets like CIFAR100 and Tiny-Imagenet datasets. We present generated synthetic images for both 1 and 10 images per class. We provide the visualization for 1 image per class on both datasets in Fig.~\ref{CIFAR100-1} and Fig.~\ref{Tiny-1},  respectively. Fig.~\ref{CIFAR100-10-1} and Fig.~\ref{CIFAR100-10-2} include visualization for CIFAR100 datasets with 10 ipc wherein each image is divided based on the number of classes. Similarly, we split the image into 50 classes for the Tiny-ImageNet dataset for 10 ipc in Fig.~\ref{Tiny-10-1} and Fig.~\ref{Tiny-10-2}.

\begin{figure*}[!t]
  \centering
   \includegraphics[keepaspectratio, width=0.8\textwidth]{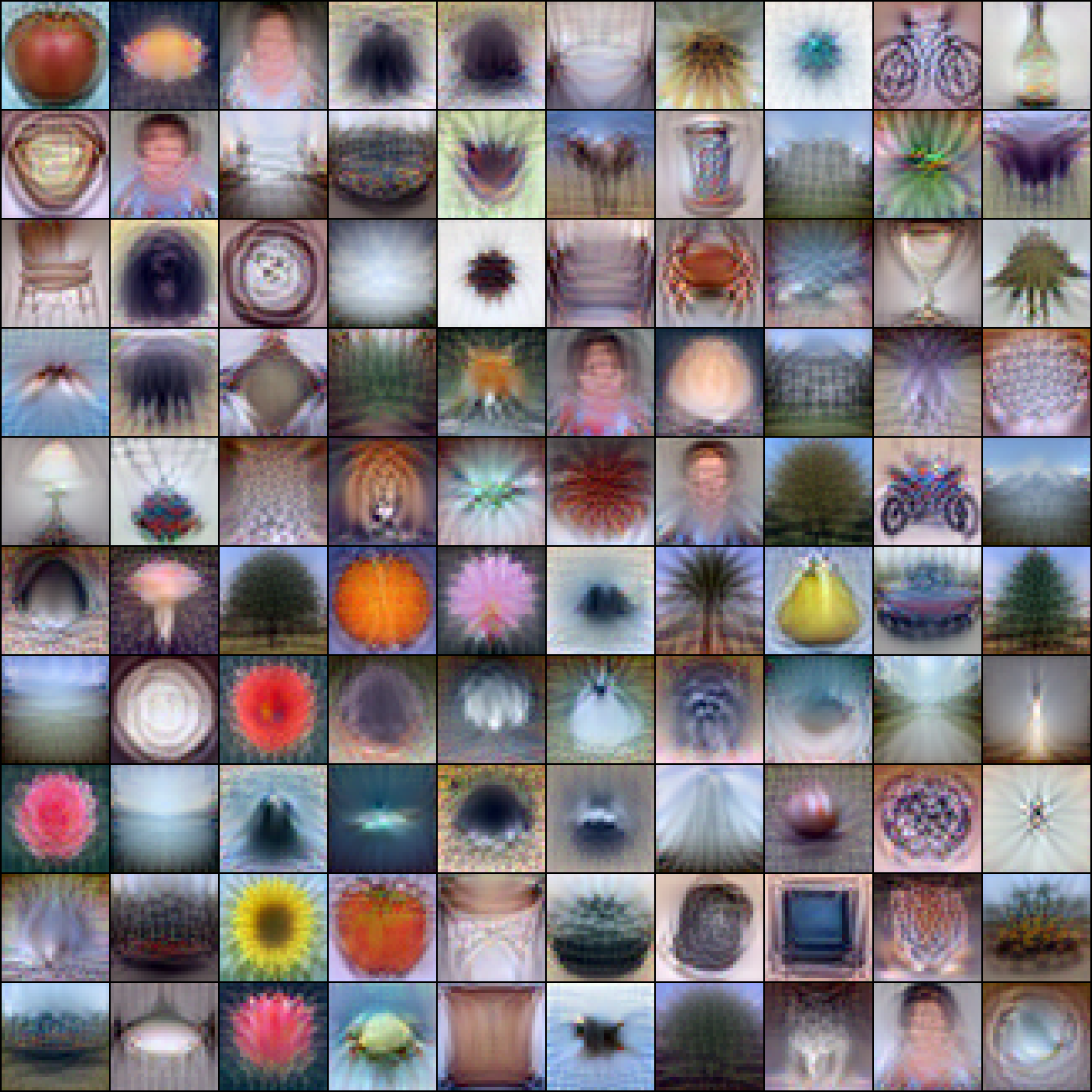}
    \caption{\label{CIFAR100-1}Visualizations of pseudo-coresets for CIFAR100 with 1 ipc.}
\end{figure*}

\begin{figure*}[!t]
   \centering
    \includegraphics[keepaspectratio, width=0.6\textwidth]{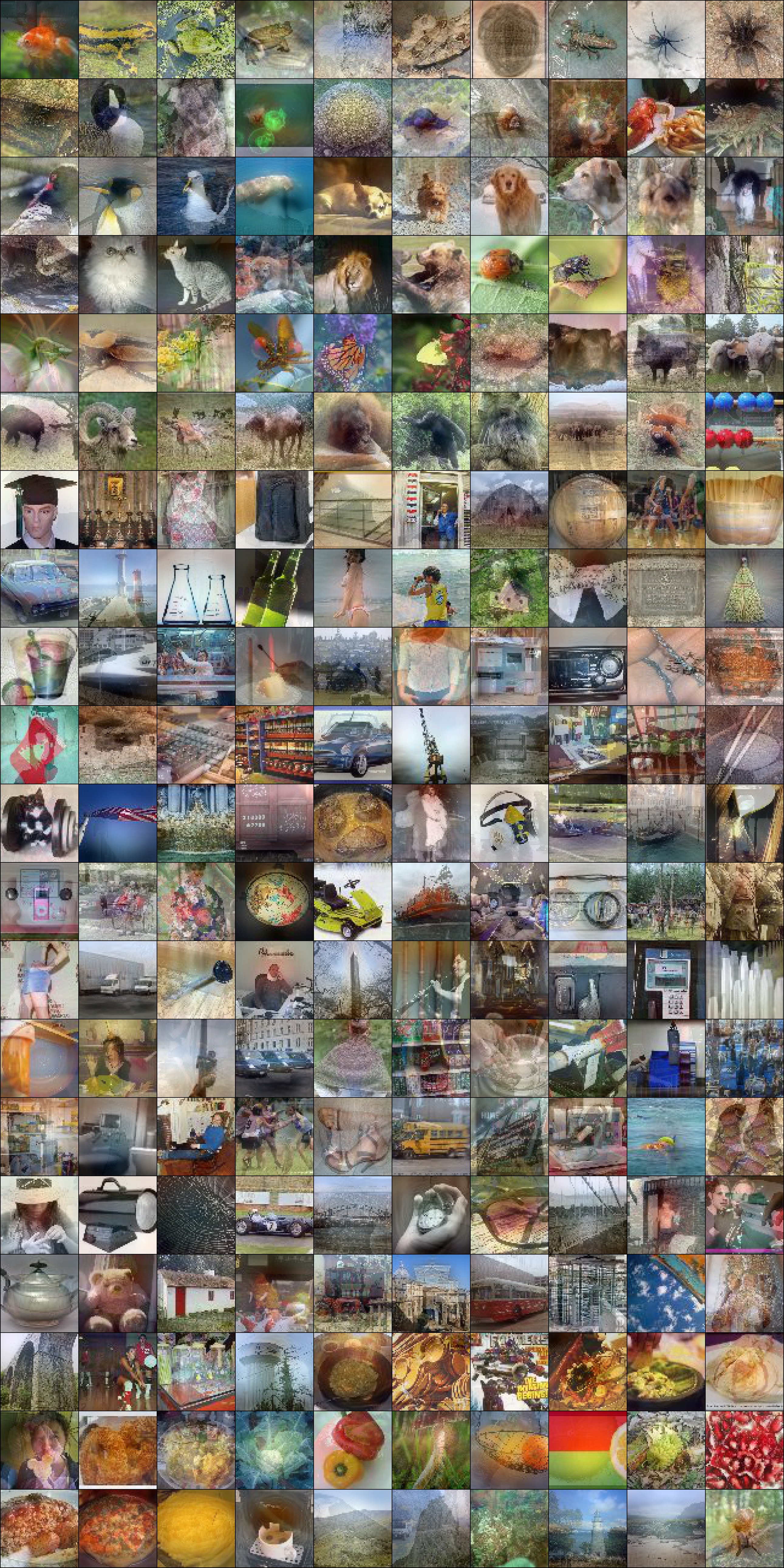}
    \caption{\label{Tiny-1}Visualizations of pseudo-coresets for Tiny ImageNet with 1 ipc.}
\end{figure*}

\begin{figure*}[!t]
   \centering
   \begin{subfigure}[t]{0.24\textwidth}
   \includegraphics[keepaspectratio,width=\textwidth]{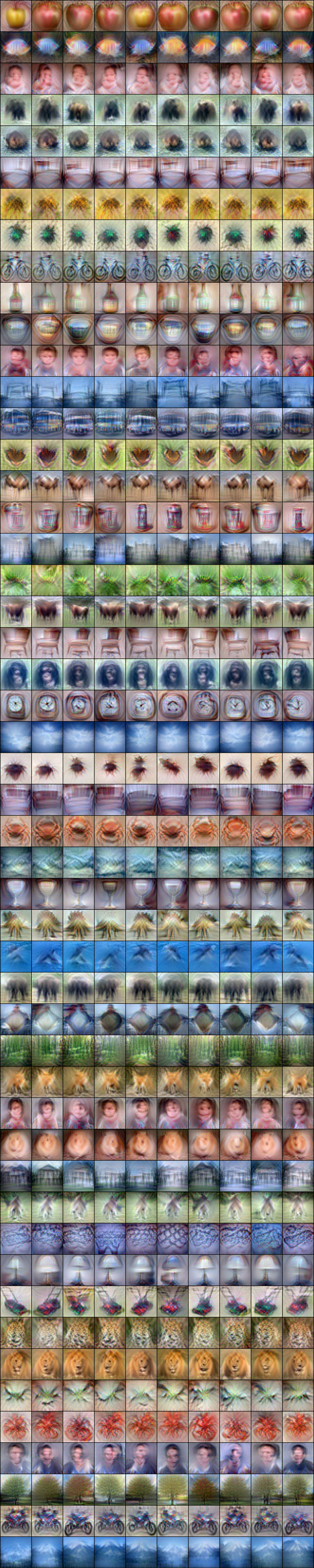}
   \caption{\label{CIFAR100-10-1}Classes 0-50}
   \end{subfigure}
   ~~~~~~~~~~~~~~~~~~~~
   \begin{subfigure}[t]{0.24\textwidth}
   \includegraphics[keepaspectratio,width=\textwidth]{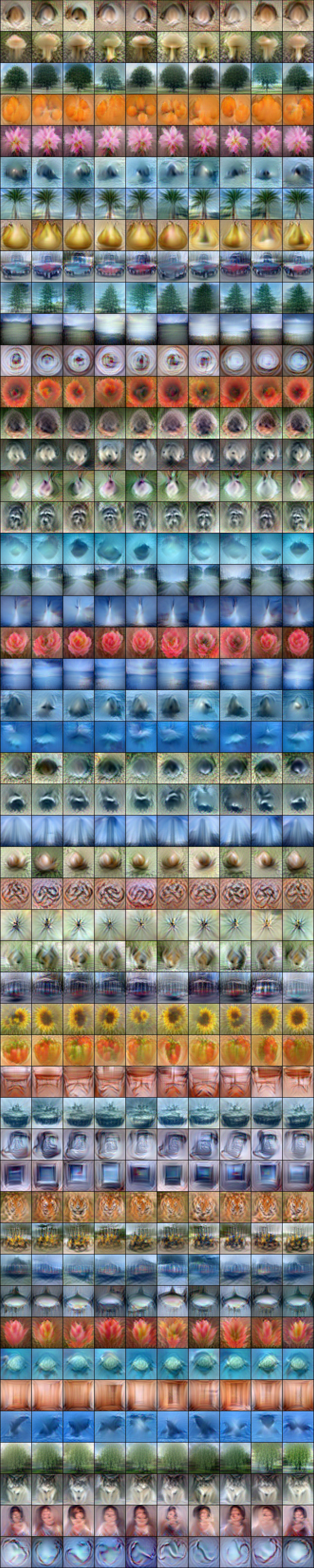}
   \caption{\label{CIFAR100-10-2}Classes 50-100}
   \end{subfigure}
   \caption{Visualizations for pseudo-coresets for CIFAR100 with 10 ipc}
\end{figure*}

\begin{figure*}[!t]
   \centering
   \begin{subfigure}[t]{0.24\textwidth}
   \includegraphics[keepaspectratio,width=\textwidth]{figs/tiny/tiny-ipc10-0-50.png}
   \caption{Classes 0-50}
   \end{subfigure}
   ~~~~~~~~~~~~~~~~~~~~
   \begin{subfigure}[t]{0.24\textwidth}
   \includegraphics[keepaspectratio,width=\textwidth]{figs/tiny/tiny-ipc10-50-100.png}
   \caption{Classes 50-100}
   \end{subfigure}
   
   \caption{\label{Tiny-10-1}Visualizations of psuedo-coresets for Tiny ImageNet with 10 ipc}
\end{figure*}

\begin{figure*}[!t]
   \centering
   \begin{subfigure}[t]{0.24\textwidth}
   \includegraphics[keepaspectratio,width=\textwidth]{figs/tiny/tiny-ipc10-100-150.png}
   \caption{Classes 100-150}
   \end{subfigure}
   ~~~~~~~~~~~~~~~~~~~~
   \begin{subfigure}[t]{0.24\textwidth}
   \includegraphics[keepaspectratio,width=\textwidth]{figs/tiny/tiny-ipc10-150-200.png}
   \caption{Classes 150-200}
   \end{subfigure}
   \caption{\label{Tiny-10-2}Visualizations of pseudo-coresets for Tiny ImageNet with  10 ipc}
\end{figure*}

\end{document}